%% file: main-long-arxiv.tex
\pdfoutput=1
\PassOptionsToPackage{table, dvipsnames, HTML}{xcolor}

\documentclass[11pt]{article}

\usepackage[preprint]{acl}
\usepackage{times}
\usepackage{latexsym}

\usepackage[T1]{fontenc}

\usepackage[utf8]{inputenc}

\usepackage{microtype}

\usepackage{inconsolata}

\usepackage{graphicx}
\usepackage{amsmath,amsfonts}
\usepackage{algorithmic}
\usepackage{algorithm}

\usepackage{microtype}
\usepackage{hyperref}
\usepackage{url}
\usepackage{booktabs}
\usepackage{graphicx}
\usepackage{cancel}
\usepackage{amsmath}
\usepackage{amssymb}
\usepackage{mathtools}
\usepackage{amsthm}
\usepackage{enumitem}
\usepackage{pifont}
\usepackage{listings}
\usepackage{fontawesome}

\usepackage{wrapfig}
\usepackage{caption}
\usepackage{mdframed}
\usepackage{multicol}
\input{math_commands}

\usepackage{booktabs}
\usepackage{multicol}
 \usepackage{multirow}

\usepackage[most]{tcolorbox}

\usepackage{listings}%
\definecolor{commentgreen}{HTML}{6A9955}
\definecolor{codepurple}{RGB}{156,39,176}

\lstdefinestyle{pytorchstyle}{
    commentstyle=\color{commentgreen}\itshape,
    keywordstyle=\color{codepurple},
    stringstyle=\color{orange},
    basicstyle=\ttfamily\footnotesize,
    breaklines=true,
    captionpos=b,
    keepspaces=true,
    showspaces=false,
    showstringspaces=false,
    showtabs=false,
    tabsize=2,
    language=Python,
    morekeywords={Tensor,Module,nn,Sequential,Linear,ReLU}
}

\title{DRA-GRPO: Your GRPO Needs to Know Diverse Reasoning Paths for Mathematical Reasoning}

\author{
\textbf{Xiwen Chen\textsuperscript{1,2*}},
\textbf{Wenhui Zhu\textsuperscript{3*}},
\textbf{Peijie Qiu\textsuperscript{4*}},
\textbf{Xuanzhao Dong\textsuperscript{3}},
\textbf{Hao Wang\textsuperscript{2}},\\
\textbf{Haiyu Wu\textsuperscript{5}},
\textbf{Huayu Li\textsuperscript{6}},
\textbf{Aristeidis Sotiras\textsuperscript{4}},
\textbf{Yalin Wang\textsuperscript{3}},
\textbf{Abolfazl Razi\textsuperscript{2}}\\
\textsuperscript{1}Morgan Stanley,
\textsuperscript{2}Clemson University,
\textsuperscript{3}Arizona State University,\\
\textsuperscript{4}Washington University in St. Louis,
\textsuperscript{5}University of Notre Dame,
\textsuperscript{6}University of Arizona
}

\begin{document}
\maketitle
\begin{abstract}

Post-training LLMs with Reinforcement Learning, specifically Group Relative Policy Optimization (GRPO), has emerged as a paradigm for enhancing mathematical reasoning. However, standard GRPO relies on scalar correctness rewards that are often \textit{non-injective} with respect to semantic content: distinct reasoning paths receive identical rewards. This leads to a \textit{Diversity-Quality Inconsistency}, where the policy collapses into a narrow set of dominant modes while ignoring equally valid but structurally novel strategies. 
To bridge this gap, we propose \textbf{D}iversity-aware \textbf{R}eward \textbf{A}djustment (\textbf{DRA}), a theoretically grounded framework that calibrates the reward signal using the semantic density of sampled groups. 
By leveraging Submodular Mutual Information (SMI), DRA implements an \textit{Inverse Propensity Scoring (IPS)} mechanism that effectively de-biases the gradient estimation. This creates a repulsive force against redundancy, driving the policy to achieve better coverage of the high-reward landscape.
Our method is plug-and-play and integrates seamlessly with GRPO variants. Empirical evaluations on five math benchmarks demonstrate that DRA-GRPO consistently outperforms strong baselines, achieving an average accuracy of 58.2\% on DeepSeek-R1-Distill-Qwen-1.5B with only 7,000 training samples and \$55 cost, highlighting the critical role of diversity calibration in data-efficient alignment. The code is available at \url{https://github.com/xiwenc1/DRA-GRPO}.
\end{abstract}

\section{Introduction}

Recent advancements in large language models (LLMs) have been significantly shaped by DeepSeek-R1-Zero~\cite{guo2025deepseekr1}, which sets a new paradigm for finetuning LLMs. Departing from traditional pipelines that rely on supervised fine-tuning (SFT) as a prerequisite, this method performs reinforcement learning (RL) directly on base models to align with human feedback. The success of DeepSeek-R1-Zero is primarily attributed to the Group Relative Policy Optimization (GRPO) algorithm~\cite{deepseekmath}, which simplifies traditional policy optimization methods for post-training~\citep{ouyang2022training}. Furthermore, studies collectively suggest that GRPO offers a more efficient and effective alternative to traditional SFT, particularly outperforming in domains requiring complex reasoning~\cite{shao2024deepseekmath,li2025reinforcement,tan2025reason}. However, while GRPO has proven highly effective, even for smaller models~\cite{dang2025reinforcement, deepscaler2025, Slow_Thinking_with_LLMs_3_Preview}, its reliance on outcome-based scalar rewards introduces a critical cognitive blind spot: the model is incentivized solely on \textit{what} the answer is, effectively blinding it to \textit{how} the reasoning path evolves.

\begin{figure}[!t]
    \centering
    \includegraphics[width=0.99\linewidth]{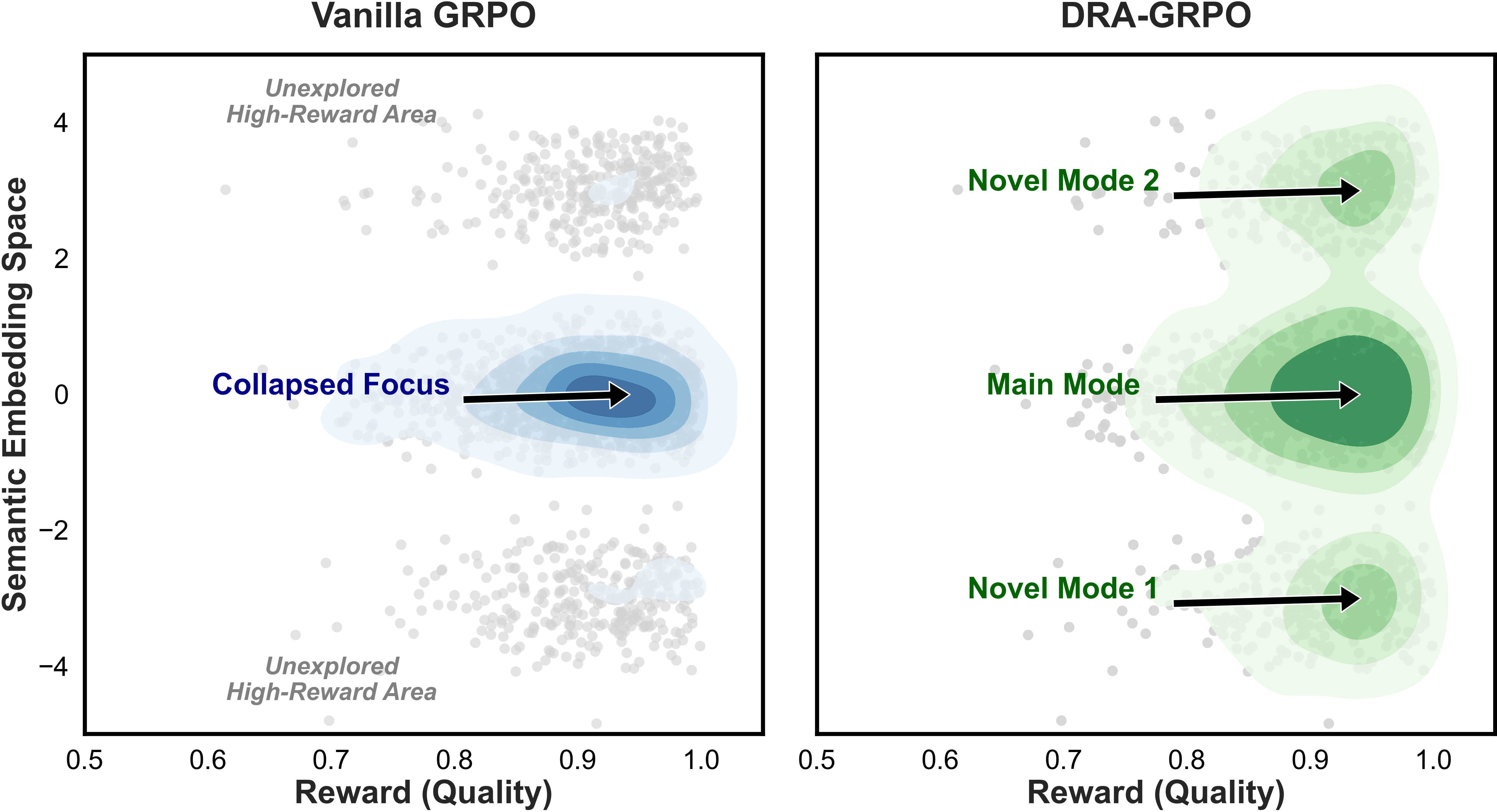}
    \caption{
\textbf{Illustration of the Exploration-Exploitation trade-off in GRPO.} 
The grey dots represent the landscape of potential high-reward reasoning paths, distributed across a common dominant mode (center) and novel but sparser modes (sides). 
\textbf{(a)} Vanilla GRPO suffers from \textit{Mode Collapse}: relying solely on scalar rewards, the policy may collapse into the dominant mode, ignoring equally valid but semantically distinct strategies.
\textbf{(b)} \textbf{DRA-GRPO} (Ours) achieves \textit{Diverse Exploration}: by penalizing semantic redundancy, our method effectively disperses probability mass to uncover and reinforce novel reasoning paths, aligning the policy with the full spectrum of correct solutions.
}
    \label{fig:teaser}
\end{figure}
This limitation leads to what we identify as the ``Diversity-Quality Inconsistency'' problem, \textit{where the scalar correctness reward fails to reflect the diverse reasoning paths leading to the same solution}. As illustrated in Fig.~\ref{fig:teaser} (\textbf{Left}), Vanilla GRPO assigns identical high rewards to all correct completions. We provide the empirical demonstration in Fig.~\ref{fig:example_case} and more discussion in Section~\ref{sec:investigation}. Consequently, despite the existence of multiple high-reward reasoning paths (represented by the grey dots), the optimization process fails to credit the \textit{diversity} of reasoning logic. This creates a biased exploration-exploitation trade-off: the model tends to collapse into a narrow set of reasoning patterns that are easiest to generate (the central dominant mode), completely neglecting the novel but sparser modes on the periphery. A concrete analogy is a teacher who assigns full credit to every student who reaches the correct answer, regardless of whether they used rote memorization or novel derivation. While outcomes are accurate, such evaluation overlooks distinct, potentially valuable reasoning strategies. This limitation is particularly critical in resource-constrained settings, where limited sampling per prompt fails to capture the full range of reasoning paths, often resulting in suboptimal policy convergence.

To bridge this gap, we propose \textit{Diversity-aware Reward Adjustment} (DRA), a plug-and-play enhancement designed to calibrate the reward signal with reasoning diversity. 
Theoretically, we frame the limitation of Vanilla GRPO as a \textit{sampling bias} problem, where the model over-samples redundant modes.
Fig.~\ref{fig:teaser}(b) demonstrates our core intuition: by integrating a diversity-aware signal, we distinguish between redundant and novel reasoning paths even when their correctness scores are identical. 
As shown in the expansive green density plot, our method successfully drives the policy to explore the peripheral ``novel modes'' that Vanilla GRPO ignores. 
Specifically, we leverage Submodular Mutual Information (SMI), instantiated with a Graph-Cut function, to implement an \textit{inverse propensity scoring} mechanism. This dynamically downweights redundant completions and amplifies the reward signal for semantically unique trajectories, effectively transforming the ``black-box'' scalar reward into a structure-aware learning signal. 
Our method integrates seamlessly with GRPO and its variant DR.~GRPO, which we refer to as \textit{DRA-GRPO} and \textit{DRA-DR.~GRPO}, respectively.

To validate the versatility and robustness of our approach, we apply DRA-GRPO across distinct model backbones, including {DeepSeek-R1-Distill-Qwen-1.5B} and {Qwen 3}. Empirical evaluations on five mathematical reasoning benchmarks demonstrate that our method is consistently effective, yielding significant performance gains over strong baselines in diverse settings. Notably, with only 7,000 training samples, our approach achieves an average accuracy of 58.2\% on the 1.5B model, confirming that explicitly modeling reasoning diversity is a fundamental key to data-efficient alignment regardless of the base model.

\section{Method}\label{sec:Method}

\subsection{Preliminary}
We briefly review the Group Relative Policy Optimization (GRPO) algorithm~\cite{deepseekmath}, as employed in~\cite{deepseekr12025}. Language model generation is formulated as a token-level Markov Decision Process (MDP).
At each generation step $t$, the state $s_t$ is the concatenation of the input question and the partial output sequence generated thus far, denoted as $s_t = \rvq; \rvo_{<t}$.
The policy $\pi_\theta(\cdot \mid s_t)$ selects the next token $o_t$ from the vocabulary $\gA$, inducing a deterministic transition to the next state $s_{t+1} = s_t; [o_t]$. Generation begins by sampling an initial state $s_1 = \rvq \sim p_{\gQ}$ from the distribution over input questions, and terminates either upon generation of the special \texttt{[eos]} token or when the token budget is exhausted. 
GRPO proposes to sample a group of responses $\mathcal{C}=\{\rvo_1, \dots, \rvo_G\}$ per question and compute their returns $\mathbf{R}=\{\{R(\rvq, \rvo_1), \dots, \{R(\rvq, \rvo_G)\}$.
Below, we present the GRPO objective, omitting the KL-divergence term for clarity.
\begin{equation}
\footnotesize
\begin{split}
    \mathcal{J}_{GRPO}&(\pi_\theta) = \E_{\rvq\sim p_{\gQ}, \{\rvo_i\}_{i=1}^G \sim \pi_{\theta_{old}}(\cdot|\rvq)} \\
    & \frac{1}{G}\sum_{i=1}^G {\frac{1}{|\rvo_i|}} \sum_{t=1}^{|\rvo_i|} \Bigg\{ \min \Bigg[ \frac{\pi_\theta(o_{i,t} | \rvq, \rvo_{i,<t})}{\pi_{\theta_{old}}(o_{i,t} | \rvq, \rvo_{i,<t})} \hat{A}_{i,t},  \\
    &\text{clip} \left( \frac{\pi_\theta(o_{i,t} | \rvq, \rvo_{i,<t})}{\pi_{\theta_{old}}(o_{i,t} | \rvq, \rvo_{i,<t})}, 1 - \epsilon, 1 + \epsilon \right)  \hat{A}_{i,t} \bigg] \Bigg\} ,
\end{split}
\label{eq:GRPO-obj}
\end{equation}
where $\hat{A}_{i,t}$ denotes the advantage function computed by:
{\footnotesize
\begin{equation}
\hat{A}_{i,t}=\frac{R(\rvq, \rvo_i) - \operatorname{mean}({ \{R(\rvq, \rvo_1), \dots, R(\rvq, \rvo_G)\} })}{ {\operatorname{std}({\{R(\rvq, \rvo_1), \dots, R(\rvq, \rvo_G)\}})}} .\label{eq:GRPO-adv}
\end{equation}}
A more recent work DR.~GRPO~\cite{liu2025understanding} proposes to remove the terms $\frac{1}{|\rvo_i|}$ and $\operatorname{std}(\cdot)$ in Eqs. \ref{eq:GRPO-obj} and \ref{eq:GRPO-adv}, to improve token efficiency. 

As our focus is on mathematical reasoning, here, we review some typical reward functions used in this task~\cite{deepseekmath,dang2025reinforcement}. Please refer to Appendix~\ref{appendix:reward} for more details.

\noindent\textbf{Accuracy Reward.} This binary metric assigns a reward of 1.0 if the parsed model output exactly matches the ground truth. 

\noindent\textbf{Cosine Reward.} To encourage conciseness alongside correctness, this function scales the reward using a cosine schedule based on completion length, assigning higher values to shorter, correct reasoning paths.

\noindent\textbf{Format Reward.} This structural constraint enforces validity by assigning a reward of 1.0 solely if the reasoning process is strictly enclosed within \texttt{<think>} and \texttt{</think>} tags.

It is worth noting that these reward functions typically compute the reward by applying straightforward criteria to the entire solution. In the following, we will argue that this would not be the optimal way to characterize the reasoning paths. 

\subsection{Diversity-Quality Inconsistency}\label{sec:investigation}

As we discussed in the previous section, both algorithms evaluate a group of independently sampled completions $\pi_{\theta_{old}}$ and reward signals typically capture only solution-level correctness, providing a sparse scalar judgment for each completion. However, this scalar reward (quality) overlooks the diverse reasoning paths that can yield identical or similar outcomes, resulting in what we term \textit{Diversity-Quality Inconsistency}. To illustrate the severity of this inconsistency, we present both a qualitative case study and a quantitative analysis.

\begin{figure*}[!t]
    \centering
    \includegraphics[width=0.99\linewidth]{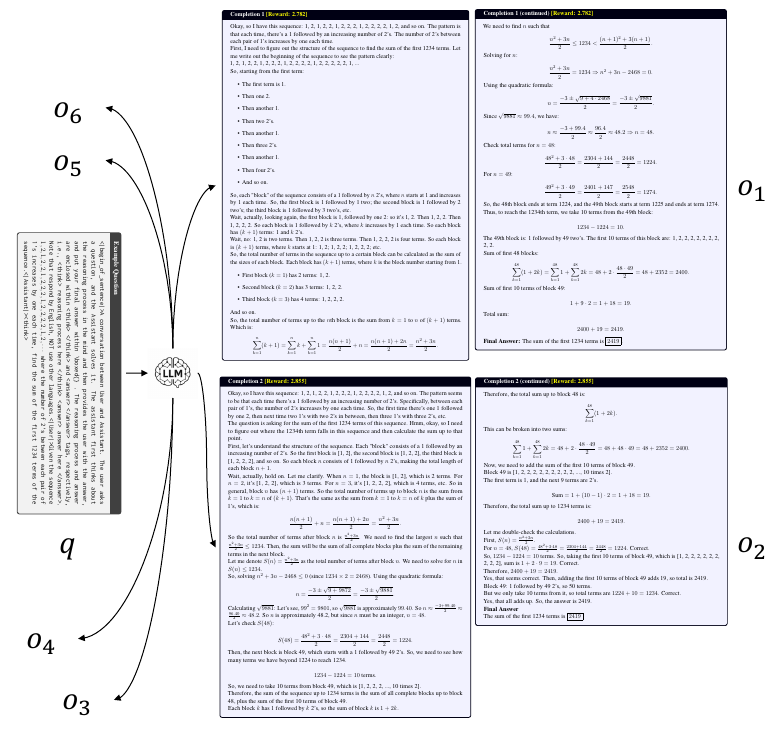}
    \caption{\textbf{Case study illustrating the \textit{Diversity-Quality Inconsistency}.} We present two correct completions for the same sequence problem. {Completion 1 ($o_1$)} adopts a concise, formula-driven strategy, whereas {Completion 2 ($o_2$)} exhibits an exploratory, ``thinking-out-loud'' reasoning style with step-by-step verification. Despite their profound semantic disparity, Vanilla GRPO assigns them nearly indistinguishable scalar rewards (2.782 vs. 2.855), failing to capture the structural diversity of the reasoning paths.}
    \label{fig:example_case}
\end{figure*}

 \noindent\textbf{Case Study.} Fig.~\ref{fig:example_case} presents two correct completions for the same sequence problem. While both trajectories converge on the correct solution, they exhibit fundamentally different cognitive structures. The first completion ($o_{1}$) adopts an exploratory, ``thinking-out-loud'' persona, characterized by a loose narrative flow and real-time self-correction mechanisms (e.g., \textit{``Wait, actually, looking again\dots''}). In stark contrast, the second completion ($o_{2}$) exhibits a systematic, didactic structure, establishing precise formal notation early and explicitly segregating the derivation from a final verification phase (e.g., \textit{``Let me double-check the calculations''}). Despite this profound semantic disparity, representing a distinction between stochastic discovery and structured verification, the scalar rewards assigned are nearly indistinguishable ($2.782$ vs.\ $2.855$). This confirms that the standard reward signal is effectively blind to the structural diversity of reasoning, treating distinct algorithmic approaches as interchangeable. More examples are shown in Appendix \ref{appendix:exmaples}.


 \noindent\textbf{Quantitative Analysis.} To investigate the relationship between reward signals and reasoning diversity, we conduct an empirical analysis over prompts with multiple sampled completions. For each prompt, we compute pairwise semantic distances between completions using cosine distance over sentence-level embeddings obtained from a pre-trained embedding model. In parallel, we compute the absolute differences in scalar reward values assigned to each completion. To measure how well reward differences reflect semantic diversity, we compute Spearman’s rank correlation coefficient between the reward distance matrix and the embedding distance matrix for each prompt.

We choose Spearman’s rank correlation for three key reasons. First, it is a \textit{non-parametric} statistic, making no assumptions about the linearity or distribution of the underlying variables, an important consideration in our setting, where reward scales and semantic distances may exhibit complex, non-linear relationships. Second, Spearman correlation is based on \textit{rank order}, allowing us to capture monotonic trends in the data, i.e., whether more semantically different completions are likely to have more divergent rewards. Third, it is \textit{robust to scale mismatches} between the two metrics (scalar rewards vs. high-dimensional embeddings), since it evaluates alignment in relative ordering rather than absolute magnitude.

\begin{figure}[!t]
    \centering
    \includegraphics[width=0.8\linewidth]{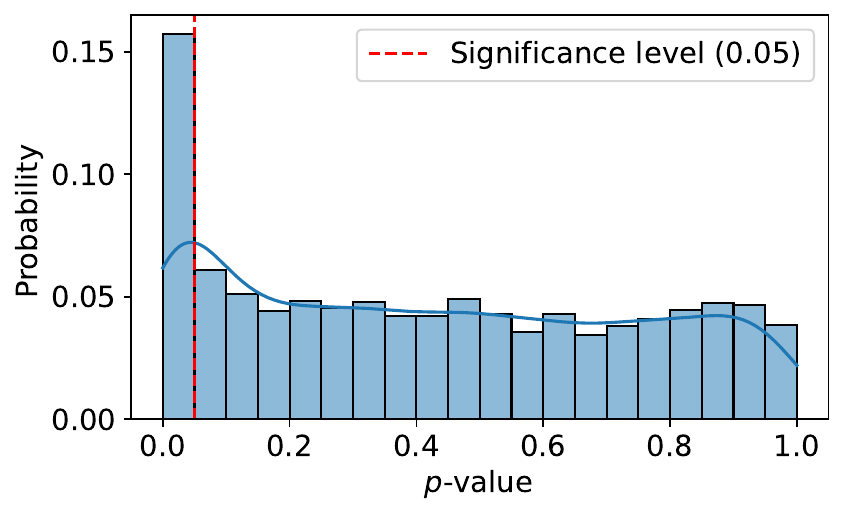}
    \caption{Distribution of $p$-values from Spearman’s rank correlation between completion quality and semantic diversity. The test is conducted for every prompt. }

    \label{fig:p1}
   
\end{figure}

We analyze the distribution of Spearman coefficients across prompts (see Fig.~\ref{fig:p1}) and observe that in the majority of cases, correlation is low or statistically insignificant ($p > 0.05$). This provides strong empirical evidence that reward alone does not capture the semantic diversity of model outputs, a phenomenon we define as the \textit{Diversity-Quality Inconsistency}. These findings motivate the need for training objectives that explicitly model and preserve reasoning diversity in addition to optimizing for correctness. Please refer to {Appendix \ref{appendix:resultInconsistency}} for more details and results for this investigation. 

\subsection{Diversity-aware Reward Adjustment} 
To address this, we propose to reweight each sample's reward based on its relative diversity/redundancy within the group: completions that are more distinct from the rest are assigned higher importance, while redundant samples are downweighted. To this end, we propose to replace $R(\rvq, \rvo_i)$ with our diversity-aware adjusted reward $\tilde{R}(\rvq, \rvo_i)$ in Eq. \ref{eq:GRPO-adv} as:
\begin{align}\label{eq:diversity-reward}
    \tilde{R}(\rvq, \rvo_i)  = \frac{R(\rvq, \rvo_i)}{1+\texttt{SMI}(\{\rvo_i\},\mathcal{C}\setminus \{\rvo_i\})},
\end{align}
where $\texttt{SMI}(\{\rvo_i\},\mathcal{C}\setminus \{\rvo_i\})$ denotes the \textit{Submodular Mutual Information} (SMI) between query completion $\rvo_i$ and the remaining completions denotes as $\mathcal{C}\setminus \{\rvo_i\}$. 
Submodular functions, with their diminishing returns property, naturally model diversity and redundancy. SMI quantifies the shared information between sets under a submodular function~\cite{iyer2021submodular,iyer2021generalized}. We instantiate SMI using the Graph-Cut function over a similarity kernel $s(\cdot, \cdot)$ presented as
\begin{align}\label{eq:SMI}
\texttt{SMI}(\{\rvo_i\}, \mathcal{C} \setminus \{\rvo_i\}) = \sum_{j \in \mathcal{C} \setminus \{\rvo_i\}} s(\rvo_i, j),
\end{align}
where we adopt the assumption that $s(\rvo_i, j)=s(j,\rvo_i)$.
It measures the total symmetric similarity between $\rvo_i$ and the remaining elements. 
In this work, we use an extra small pretrained model to get the embedding for each completion. 
Due to submodularity, this formulation captures diminishing redundancy: elements more similar to the set contribute less marginal information. Thus, Graph-Cut SMI provides a principled measure of $\rvo_i$'s relative redundancy (high value) or diversity (low value) within the group. In the context of reward adjustment in Eq. \ref{eq:diversity-reward}, we assign a more redundant completion with a lower weight to its corresponding reward and a diverse completion a higher weight. We use cosine similarity as the kernel $s(\cdot)$, enabling efficient computation of the SMI via a precomputed similarity matrix. This can be presented as
\begin{align}
\small
    \tilde{R}(\rvq, \rvo_i)  &= \frac{R(\rvq, \rvo_i)}{1+\texttt{SMI}(\{\rvo_i\},\mathcal{C}\setminus \{\rvo_i\})} \\ \nonumber
    &= \frac{R(\rvq, \rvo_i)}{1+\sum_{j \in \mathcal{C} \setminus \{\rvo_i\}} s(\rvo_i, j)} \\ \nonumber
    &= \frac{R(\rvq, \rvo_i)}{s(\rvo_i, \rvo_i)+\sum_{j \in \mathcal{C} \setminus \{\rvo_i\}} s(\rvo_i, j)} \\ \nonumber
    & = \frac{R(\rvq, \rvo_i)}{\sum_{j=0}^G \boldsymbol{L}_{ij} }.
\end{align}
We note that $\sum_{j=0}^G \boldsymbol{L}_{ij}$ is the sum of the $i$th row of the similarity matrix $\boldsymbol{L}$, so this can be efficiently computed through \texttt{Pytorch} tensor operation trick for all completions as shown in Algorithm \ref{alg:000}, i.e., \texttt{similarity\_matrix.sum(dim=1)}.
This results in a total computational complexity of $\mathcal{O}(G^2)$ for a group of size $G$. We provide a PyTorch-style algorithmic summary in Algorithm \ref{alg:000} (Appendix \ref{appendix:Algorithmic}).

\subsection{Theoretical Justification}
\label{sec:probabilistic_view}

To theoretically ground our approach, we analyze the SMI-based adjustment through the lens of importance sampling and density estimation.
Let $\mathcal{Z}$ denote the latent semantic space. Standard GRPO approximates gradients using samples drawn from a proposal distribution $q(\rvo) \triangleq \pi_{\theta_{\text{old}}}(\rvo|\rvq)$. 
This standard objective effectively maximizes the reward weighted by the sampling prior:
\begin{equation}
\small
    \nabla \mathcal{J}_{\text{GRPO}} \approx \mathbb{E}_{\rvo \sim q} \left[ R(\rvo) \nabla \log \pi \right] = \int_{\mathcal{Z}} q(\rvo)R(\rvo) \nabla \log \pi \, d\rvo.
\end{equation}
However, $q(\rvo)$ is inherently biased: the model concentrates probability mass on ``dominant modes'' (easy-to-generate patterns), leading to the over-sampling of redundant paths. Our proposed SMI term functions as a \textit{Kernel Density Estimator (KDE)} of this biased proposal distribution, i.e., $\hat{p}_G(\rvo) \approx q(\rvo)$. Under this framework, the adjusted reward $\tilde{R}$ implements \textit{Inverse Propensity Scoring (IPS)}~\cite{rosenbaum1983central}. By scaling the raw reward $R$ by the inverse of the estimated density, we reweight the gradient estimate:
\begin{equation}
\small
\begin{split}
    \nabla \mathcal{J}_{\text{DRA}} &\approx \mathbb{E}_{\rvo \sim q} \left[ \frac{R(\rvo)}{\hat{p}_G(\rvo)} \nabla \log \pi \right] \\
    &\approx \int_{\mathcal{Z}} q(\rvo) \frac{R(\rvo)}{q(\rvo)} \nabla \log \pi \, d\rvo = \int_{\mathcal{Z}} R(\rvo) \nabla \log \pi \, d\rvo.
\end{split}
\end{equation}
Critically, the Right-Hand Side (RHS) represents the gradient over the \textit{true reward landscape}, independent of the sampling bias $q(\rvo)$.
In contrast to standard GRPO, which is heavily weighted by the model's prior $q$, our objective approximates $\int R(\rvo)d\rvo$. This implies that \textbf{all} high-reward regions, whether they are dominant modes or rare novel paths (as shown in Fig.~\ref{fig:teaser}), contribute approximately equally to the optimization, thereby encouraging \textit{better coverage} of the diverse solution space.


\section{Experiment}
\subsection{Experimental Setup}
\noindent\textbf{Training Dataset.} We adopt a high-quality dataset curated by \citep{dang2025reinforcement}. This dataset consists of only \textbf{7000} samples refined and selected from the s1 dataset~\citep{s1simpletesttimescaling} and the DeepScaleR dataset~\citep{deepscaler2025} with mixed problem difficulties. 

\noindent\textbf{Evaluation Dataset.} 
We select five popular mathematical reasoning benchmarks: AIME24~\footnote{\url{https://huggingface.co/datasets/AI-MO/aimo-validation-aime}}, MATH-500~\citeyearpar{lightman2023lets,hendrycksmath2021}, AMC23~\footnote{\url{https://huggingface.co/datasets/AI-MO/aimo-validation-amc}}, Minerva~\citeyearpar{minervamath} and OlympiadBench~\citeyearpar{he-etal-2024-olympiadbench}.

\noindent\textbf{Baselines.} We evaluate our approach against various baseline models. The general-purpose large model: (i) \texttt{Llama-3.1-70B-Instruct}~\citep{MetaAI2024} and (ii) \texttt{o1-preview}~\citep{o1preview}.
For mathematics-focused 7B models, we consider: (iii) \texttt{Qwen-2.5-Math-7B-Instruct}~\citep{yang2024qwen25mathtechnicalreportmathematical}; (iv) \texttt{rStar-Math-7B}~\citep{guan2025rstar}; (v) \texttt{Eurus-2-7B-PRIME}~\citep{cui2025processreinforcementimplicitrewards}; and (vi) \texttt{Qwen2.5-7B-SimpleRL}~\citep{zeng2025simplerl}. Lastly, for mathematics-focused 1.5B models, instead of our base model, we include (vii) \texttt{DeepScaleR-1.5B-Preview}~\citep{deepscaler2025}, (viii) \texttt{Still-3-1.5B-Preview}~\citep{Slow_Thinking_with_LLMs_3_Preview}, and (ix) Open-RS~\cite{dang2025reinforcement}.

\input{table1}

\noindent\textbf{Implementation.} 
Following the experimental protocols of Open-RS~\cite{dang2025reinforcement}, we adopt \texttt{DeepSeek-R1-Distill-Qwen-1.5B}~\cite{deepseekr12025} as our primary base model to ensure a fair and direct comparison with state-of-the-art baselines. 
We use 4x NVIDIA A100 40GB GPUs. Please refer to Appendix~\ref{appendix:implement} for the details of hyperparameters.
Additionally, to demonstrate the generalization of our method to newer architectures, we provide ablation studies using \texttt{Qwen3-4B-Instruct} in the following subsection.

\subsection{Empirical Analysis}\label{sec:Empirical}

\noindent\textbf{Main Results in Accuracy.} As shown in Table~\ref{tab:main_result}, our primary observation is that integrating our method with DR.~GRPO outperforms all baseline approaches across various parameter scales, achieving an average accuracy of 58.2\% across all benchmarks. Notably, it achieves the highest accuracy on both AMC23 (85\%) and OlympiadBench (53.8\%). When incorporated with GRPO, our method obtains an average accuracy of 56.7\%, which is on par with the previous state-of-the-art, \texttt{DeepScaleR-1.5B-Preview} (57\%). However, our approach requires only 7,000 fine-tuning samples, in contrast to the approximately 40,000 samples used by \texttt{DeepScaleR-1.5B-Preview}. These results demonstrate the superiority of our method in low-resource settings, i.e., a small model with 1.5B parameters and limited samples for fine-tuning.

\noindent\textbf{Ablation on Algorithms.} 
The ablation results are summarized in Table~\ref{tab:main_result}. The main observation is that, compared to the base model \texttt{DeepSeek-R1-Distill-Qwen-1.5B}, our methods yield improvements of 7.8\% and 9.3\% in average accuracy. More importantly, integrating our method with GRPO leads to a 1.9\% increase in accuracy compared to using GRPO alone. A similar conclusion can be drawn for DR.~GRPO, where our method achieves an average accuracy gain of 2.2\% across all benchmarks. We also highlight several notable improvements: our method boosts performance on AIME24 by 6.7\% and 3.4\% for GRPO and DR.~GRPO, respectively, and achieves a 5\% gain on AMC23 with DR.~GRPO. These results further confirm the effectiveness of our method.

While our main experiments focus on \texttt{DeepSeek-R1-Distill-Qwen-1.5B} to align with the Open-RS benchmark, it is crucial to verify if our method generalizes to stronger, state-of-the-art models. To this end, we conducted additional ablation studies on \texttt{Qwen3-4B-Instruct}, as summarized in Table~\ref{tab:aba_basemodel1}. Consistent with our observations on the 1.5B model, DRA brings stable improvements across different architectures. Specifically, when integrated with DR.~GRPO, our method improves the average accuracy from 73.3\% to 74.1\% ($\Delta+0.8\%$). This confirms that the benefits of explicitly modeling semantic diversity are not limited to specific parameter scales and remain effective even with stronger base models.

\begin{table}[h]
\centering
\caption{Ablation analysis on different base models. Here we use \texttt{Qwen3-4B-Instruct}~\cite{qwen3technicalreport}.}
\vspace{-0.1in}
\label{tab:aba_basemodel1}
\resizebox{0.3\textwidth}{!}{%
\begin{tabular}{lcccc}
\toprule
\multicolumn{1}{c}{\textbf{Model}} & \multicolumn{1}{c}{\textbf{Baseline}} & \multicolumn{1}{c}{\textbf{+DRA}} & \multicolumn{1}{c}{\textbf{$\Delta$}} \\ 
\midrule
GRPO     & 73.4 & 74.3 & +0.9 \\ 
DR. GRPO & 73.3 & 74.1 & +0.8 \\ 
\bottomrule
\end{tabular}%
}
\end{table}

\noindent\textbf{Ablation on Embeddings.}
To investigate the sensitivity of our method to the choice of semantic representations, we replace the default embedding model (\texttt{jina-embeddings-v2-small-en}) with \texttt{nomic-embed-text-v1.5}~\cite{nussbaum2025nomic}. As shown in Table~\ref{tab:aba_embedding}, our method maintains consistent performance gains, improving GRPO by 1.5\% and DR.~GRPO by 1.6\%. Although the absolute performance is slightly different from the main results, the stable improvements confirm that \textit{DRA-GRPO} is robust to different embedding spaces and effectively captures semantic diversity regardless of the underlying encoder.

\begin{table}[h] 
\centering
\caption{Ablation analysis on different embedding models. Here we use \texttt{nomic-embed-text-v1.5}~\cite{nussbaum2025nomic}.}
\vspace{-0.1in}
\label{tab:aba_embedding} 
\resizebox{0.4\textwidth}{!}{%
\begin{tabular}{lcccc}
\toprule
\multicolumn{1}{c}{\textbf{Model}} & \multicolumn{1}{c}{\textbf{Baseline}} & \multicolumn{1}{c}{\textbf{+DRA (Nomic)}} & \multicolumn{1}{c}{\textbf{$\Delta$}} \\ 
\midrule
GRPO     & 54.8 & \textbf{56.3} & \textbf{+1.5} \\
DR. GRPO & 56.0 & \textbf{57.6} & \textbf{+1.6} \\ 
\bottomrule
\end{tabular}%
}
\end{table}

\noindent\textbf{Efficiency.} Compared to the vanilla GRPO and DR.~GRPO, our method introduces a small overhead due to encoding the completions. As shown in the table following, our method introduces approximately 6\% runtime and 1.4\% GPU overhead. 
\begin{wraptable}[4]{r}{0.6\linewidth}
\centering
\label{tab:overhead}
\resizebox{0.99\linewidth}{!}{%
\begin{tabular}{lccc}\toprule
 & Runtime & GPU\\ \midrule
Vanilla & $\sim$84s/step & $\sim$38.77GB/device  \\
+ DGA & $\sim$90s/step & $\sim$39.34GB/device \\ \bottomrule
\end{tabular}%
}
\end{wraptable}
This result suggests that the introduced overhead is relatively minor and unlikely to impact practical deployment.

\noindent\textbf{Comparison with Other SMI.} 
Another potential diversity-based SMI is known as logdet SMI~\cite{iyer2021generalized}. This is related to the quality-diversity decomposition in determinantal point processing (DPP)~\cite{kulesza2012determinantal}. 
In our context, it is defined as
\begin{align}
&\texttt{SMI}(\{\rvo_i\}, \mathcal{C} \setminus \{\rvo_i\}) \\ \nonumber
=& \log\det \boldsymbol{L}_{ii}+\log\det \boldsymbol{L}_{\mathcal{C} \setminus \{\rvo_i\}} - \log\det \boldsymbol{L}_{\mathcal{C}},
\end{align}\label{eq:logdetSMI}
where $\boldsymbol{L}_{ii}=1$ denotes the $i$th diagonal value of the similarity matrix, and its value is 1 as we use a cosine similarity kernel. $\boldsymbol{L}_{\mathcal{C} \setminus \{\rvo_i\}}$ and $\boldsymbol{L}_{\mathcal{C}}$ denote the rows and columns indexed by the set $\mathcal{C} \setminus \{\rvo_i\}$ and $\mathcal{C}$, respectively. Although we need a complexity of $\mathcal{O}(G^3)$ to precompute $\log\det \boldsymbol{L}_{\mathcal{C}}$, for each $\rvo_i$, we need to compute $\log\det\boldsymbol{L}_{\mathcal{C} \setminus \{\rvo_i\}}$, which is obviously less efficient than Graph-cut SMI and would be challenging for scaling.

To validate this, Table~\ref{tab:aba_smi} compares Logdet SMI against our Graph-Cut SMI. In terms of accuracy, both methods yield comparable gains, confirming the robustness of penalizing redundancy. However, the key distinction lies in scalability. Logdet is over $35\times$ more expensive than our method ($\approx$ 1573.93 vs. 44.15 $\mu$s/prompt). Furthermore, as group size $G$ scales, Logdet's $\mathcal{O}(G^3)$ complexity becomes a computational bottleneck. In contrast, our $\mathcal{O}(G^2)$ approach remains lightweight. Thus, our method offers a superior balance between performance and efficiency.

\begin{table}[h]
\centering
\caption{Ablation analysis on different SMI functions.}
\label{tab:aba_smi}
\resizebox{0.48\textwidth}{!}{%
\begin{tabular}{lcccc}
\toprule
\multicolumn{1}{c}{\textbf{Model}} & \multicolumn{1}{c}{\textbf{Baseline}} & \multicolumn{1}{c}{\textbf{Logdet}} & \multicolumn{1}{c}{\textbf{Ours}} \\ 
\midrule
GRPO     & 54.8 & 56.8 & 56.7 \\
DR. GRPO & 56.0 & 58.0 & \textbf{58.2} \\ 
\midrule
Runtime (Compute weight) & -- & 1573.93 $\mu$s/prompt & \textbf{44.15 $\mu$s/prompt} \\ 
\bottomrule
\end{tabular}%
}
\end{table}

\noindent\textbf{Training Cost.} Training for 500 steps takes approximately 12.5 hours on a 4\texttimes A100 setup, costing an estimated \$55, which is on par with Open-RS~(\citeyear{dang2025reinforcement}). 
Table \ref{tab:cost} provides more comparisons with different methods.
\input{table2}

\section{Discussion}

\noindent\underline{\textit{Calibrated Exploration vs. Biased Exploitation.}} 
Standard GRPO faces a critical trade-off: it exploits via scalar rewards but explores heavily based on the model's prior, often leading to over-exploitation of ``dominant modes.'' 
Our theoretical framing of \textit{Inverse Propensity Scoring} (IPS) explains how DRA resolves this. 
Instead of relying on random noise (e.g., high temperature) for exploration, DRA introduces a structured ``repulsive force'' in the optimization landscape. 
By down-weighting redundant samples (high propensity), we mathematically force the policy to shift its probability mass toward the low-density, high-reward regions. 
This transforms exploration from a stochastic process into a \textit{calibrated} one, ensuring that the model explores semantically distinct paths proportional to their uniqueness rather than their frequency.

\vspace{0.5em}

\noindent\underline{\textit{Ad-hoc vs. Post-hoc Diversity Modeling.}} 
Existing strategies for diversity typically fall into two categories: \textit{ad-hoc} and \textit{post-hoc}. 
Ad-hoc approaches, such as tuning sampling temperature, attempt to encourage diversity during generation but fail to explicitly model correlations among completions. 
Conversely, post-hoc selection methods like Determinantal Point Processes (DPPs) filter a large pool of samples to find a diverse subset~\cite{kulesza2012determinantal}. 
While effective for inference, DPPs are data-inefficient for training because they discard valid samples, reducing the effective batch size for gradient updates.
Our SMI-based approach represents a superior \textit{integration strategy}: it preserves all sampled data but dynamically adjusts their importance weights. 
This allows the model to learn from the entire distribution, learning \textit{what not to repeat} from redundant samples and \textit{what to reinforce} from novel ones, without the sample waste associated with filtering.

\section{Related Work}

LLM alignment has transitioned from traditional PPO-based RLHF~\cite{schulman2017proximal, ouyang2022training} to the ``R1-Zero'' paradigm~\cite{guo2025deepseekr1}. 
GRPO~\cite{deepseekmath} facilitates this transition by replacing the resource-intensive critic network with group-relative advantage estimation. 
Since its inception, several variants have emerged to improve this framework. 
For example, DR.~GRPO~\cite{liu2025understanding} focuses on unbiased advantage estimation and token-level efficiency, while DAPO~\cite{yu2025dapo} introduces a ``Clip-Higher'' mechanism to mitigate entropy collapse by relaxing optimization constraints. 
Other explorations like Info-GRPO~\cite{anonymous2025infogrpo} and Critique-GRPO~\cite{zhang2025critique} incorporate mutual information maximization or natural language feedback to refine the reasoning process.

However, existing optimization-centric methods like DAPO and Info-GRPO still rely on uncalibrated scalar rewards that neglect \textit{semantic density}. 
While DAPO refines learning dynamics via relaxed clipping and Info-GRPO promotes diversity implicitly by conditioning on input latent priors (e.g., random seeds), they primarily operate on optimization constraints or input perturbations. 
In contrast, \textit{DRA-GRPO} represents a significant departure by addressing the root cause: \textit{sampling bias} in the reward signal.
By intervening directly at the reward level to penalize semantic redundancy, our method effectively \textit{calibrates} the exploration pressure using the density of the generated distribution. 
This makes our method a plug-and-play enhancement for the GRPO family that works independently of gradient adjustments.

We choose DR.~GRPO (in addition to vanilla GRPO) as our primary baseline because it represents the state-of-the-art in unbiased GRPO training. 
We do not include direct comparisons with variants like Critique-GRPO~\cite{zhang2025critique} because they rely on additional supervision signals, specifically natural language critiques, whereas our method focuses purely on the internal scalar reward dynamics.

\section{Conclusion}

In this paper, we introduced \textit{DRA-GRPO}, a diversity-aware reward adjustment mechanism designed to resolve the ``Diversity-Quality Inconsistency'' in R1-Zero-like training. 
Unlike previous optimization-centric variants, our approach targets the semantic informativeness of the reinforcement signal itself. 
We provide a theoretical justification for our method, demonstrating that penalizing redundancy via Submodular Mutual Information (SMI) can be viewed as \textit{Inverse Propensity Scoring} (IPS).
This effectively de-biases the gradient estimation, allowing the policy to explore the full support of the high-reward landscape rather than collapsing into the model's prior.

Our empirical results confirm that \textit{DRA-GRPO} yields significant performance gains over state-of-the-art baselines like DR.~GRPO, particularly in scenarios with constrained sampling budgets. 
The modular and plug-and-play nature of our method makes it a versatile enhancement for the evolving GRPO framework. 
Ultimately, this work highlights the necessity of calibrating scalar rewards with semantic density, providing a rigorous probabilistic foundation for fostering robust reasoning capabilities in large language models.



\section{Limitations}

\begin{itemize}
    \item \textbf{Community-Wide Neglect of Reasoning Path Verification:} The prevailing research ecosystem predominantly prioritizes final-answer accuracy, resulting in a systematic oversight regarding the logical validity of intermediate reasoning steps. Existing benchmarks and automated evaluation protocols are designed to reward outcome matching, largely ignoring whether the specific reasoning paths taken are intrinsically correct. This field-wide gap means that robust, scalable methodologies for assessing the correctness of diverse reasoning traces remain undeveloped, leaving the rigorous validation of fine-grained reasoning quality as an unresolved challenge for the broader community.
    
\end{itemize}
   

\section*{Acknowledgments}
\label{sec:Acknowledgement}

This material is based upon the work supported by the National Science Foundation under Grant Number CNS-2204721.

\bibliography{custom}

\clearpage
\appendix

\renewcommand{\thefigure}{S\arabic{figure}}
\renewcommand{\thetable}{S\arabic{table}}
\section{Reward Function in Mathematical Reasoning}\label{appendix:reward}

We show some typical reward functions below. These functions often compute the reward based on some simple rules, which fail to explicitly capture the inherent semantic diversity among completions. 

\noindent\textbf{Accuracy Reward.} This function assigns binary rewards to model completions based on exact agreement with the ground truth solution. It begins by parsing the ground truth using a LaTeX extraction configuration and skips evaluation with a full reward of 1.0 if the solution is unparseable. For valid cases, the model’s output is also parsed with normalization settings that enforce clean LaTeX formatting, including handling of boxed expressions and units. The parsed output is compared against the ground truth using a verification function. If they match exactly, the function assigns a reward of 1.0; otherwise, the reward is 0.0.

\noindent\textbf{Cosine (Correctness) Reward.} This is an upgraded version of Accuracy Reward. It computes rewards for model completions by evaluating their correctness and scaling the reward based on completion length using a cosine schedule. For each completion, it parses both the model output and the ground truth solution using a LaTeX-aware parsing configuration. If parsing fails for the ground truth, the function assigns a default reward of 1.0 and skips evaluation. Correctness is verified by comparing the parsed outputs. The reward is then determined by a cosine function of the output length relative to a maximum length parameter, encouraging shorter correct answers by assigning them higher rewards and penalizing shorter incorrect ones more heavily. 

\noindent\textbf{Format Reward.} This function is designed to evaluate a list of completions by checking whether the reasoning process is properly enclosed within \texttt{<think>} and \texttt{</think>} tags. It defines an internal function \texttt{count\_tags} that inspects each text for exactly one occurrence of the \texttt{\textbackslash n</think>\textbackslash n} tag sequence. This is because the opening \texttt{<think>} tag is assumed to be present in the system prompt and thus does not need to be counted. The function extracts the content strings from the completions, applies the \texttt{count\_tags} function to each, and returns a list of floating-point scores. A score of 1.0 is assigned if the proper \texttt{</think>} tag format is found exactly once; otherwise, a score of 0.0 is given. 

\section{Investigation on Diversity-Quality Inconsistency}\label{appendix:resultInconsistency}

\begin{figure}
    \centering
    \includegraphics[width=0.99\linewidth]{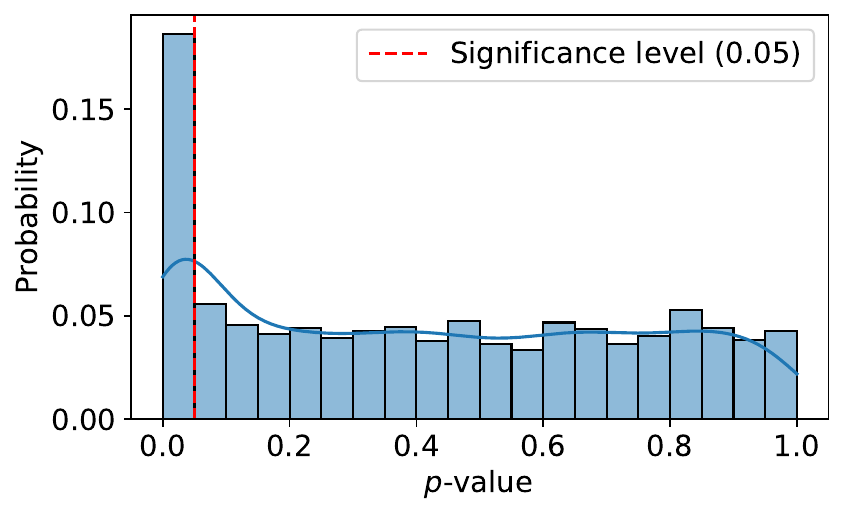}
    \caption{Distribution of $p$-values from Spearman’s rank correlation between completion quality and semantic diversity. Embedding model is \texttt{nomic-ai/nomic-embed-text-v1.5}.}

    \label{fig:p2}
   
\end{figure}

We also show a result by using a different embedding model \texttt{nomic-ai/nomic-embed-text-v1.5}~\cite{nussbaum2025nomic} in Fig. \ref{fig:p2}. Similarly, for over 80\% prompts, their completion diversity and rewards are irrelevant.

\section{Algorithmic Summary}\label{appendix:Algorithmic}

Please refer to Algorithm \ref{alg:000}.
\input{algorithm}

\section{Implementation Detail}\label{appendix:implement}
We provide our hyperparameters for both  GRPO and DR. GRPO is in the table below \ref{tab:hypers}. The implementation is based on the source code of \texttt{trl} package from Huggingface~\cite{vonwerra2022trl}. The training pipeline and prompt setups are based on \url{https://github.com/knoveleng/open-rs}.   
We carefully select a small model, \texttt{jina-embeddings-v2-small-en}~\cite{günther2023jina}, as the completion embedding model, which supports processing a sequence with up to 8192 tokens. The reason is that we want to preserve the efficiency, and we do not tend to adjust original hyperparameters, such as mini-batch size.

\begin{table}[htbp]
\centering
\caption{Hyperparameter Setups for our trainers.}
\resizebox{0.47\textwidth}{!}{
\begin{tabular}{lc}
\toprule
\textbf{Parameter} & \textbf{Value} \\
\midrule
\multicolumn{2}{l}{\textit{General Settings}} \\
bf16 & true \\
use\_vllm & true \\
vllm\_device & auto \\
vllm\_enforce\_eager & true \\
vllm\_gpu\_memory\_utilization & 0.7 \\
vllm\_max\_model\_len & 4608 \\
do\_eval & false \\
\midrule
\multicolumn{2}{l}{\textit{Training Configuration}} \\
gradient\_accumulation\_steps & 4 \\
gradient\_checkpointing & true \\
gradient\_checkpointing\_kwargs & use\_reentrant: false \\
learning\_rate & 1.0e-06 \\
lr\_scheduler\_type & cosine\_with\_min\_lr \\
lr\_scheduler\_kwargs & min\_lr\_rate: 0.1 \\
warmup\_ratio & 0.1 \\
max\_steps & 500 \\
num\_train\_epochs & 1 \\
per\_device\_train\_batch\_size & 6 \\
per\_device\_eval\_batch\_size & 6 \\
\midrule
\multicolumn{2}{l}{\textit{Generation Settings}} \\
max\_prompt\_length & 512 \\
max\_completion\_length & 3584 \\
num\_generations & 6 \\
temperature & 0.7 \\
\midrule
\multicolumn{2}{l}{\textit{Reward Configuration}} \\
reward\_funcs & format, accuracy (cosine) \\
reward\_weights & 1.0, 2.0 \\
\bottomrule
\end{tabular}
}
\label{tab:hypers}
\end{table}

\section{Case Study: Examples of Diverse Completions}\label{appendix:exmaples}

Here, we present selected examples from the GRPO training process to illustrate the key motivation of our paper.
Given the same problem, the LLM can generate diverse answers; however, these answers often receive very similar reward scores. This suggests that learning based on solution-level judgments may fail to distinguish between different reasoning paths.
Below, we show two cases that produce correct answers but demonstrate distinct reasoning perspectives and styles. We also present an example where both completions follow coherent reasoning processes but result in incorrect answers.

\subsection{Example 1} 

\noindent\textbf{Question:} Fig. \ref{q1}. 

\noindent\textbf{Two Completions:} (i) Fig. \ref{q1a1} and (ii) Fig. \ref{q1a2}.

\noindent\textbf{Short Analysis.}
While both outputs correctly arrive at the answer \( \boxed{1007} \), they reflect notably different problem-solving \textbf{perspectives}.

The first response adopts an empirical, trial-based strategy. Its reward score is 2.103. The model explores specific candidate values of the divisor \( m \), such as 1007, 1008, and 1009, and evaluates the resulting remainders. This process mimics a human-like, exploratory reasoning pattern, i.e., tentative, iterative, and conversational—ultimately identifying that \( m = 1008 \) yields the maximum remainder \( 1007 \). The approach is grounded in pattern recognition and error correction, reflecting a ``numerical experimentation'' mindset often used by learners.

In contrast, the second response applies a more principled, algebraic perspective. Its reward score is 2.110, almost the same as the first one. The model leverages the mathematical identity that the maximum remainder when dividing \( a \) by \( m \) is \( m - 1 \), which occurs when \( a \equiv -1 \mod m \), or equivalently, when \( m \mid (a + 1) \). Using this, it reduces the problem to finding the largest proper divisor of 2016. It proceeds to factor 2016 as \( 2^5 \times 3^2 \times 7 \) and identifies \( m = 1008 \) as the largest valid divisor, yielding \( n = 1007 \). This response demonstrates structured mathematical reasoning and modular arithmetic awareness, providing a generalizable method beyond this specific example.



\subsection{Example 2} 

\noindent\textbf{Question:} Fig. \ref{q2}. 

\noindent\textbf{Two Completions:} (i) Figs. \ref{q2a1-a} and \ref{q2a1-b} and (ii) Figs. \ref{q2a2-a} and \ref{q2a2-b}.

\noindent\textbf{Short Analysis.}
Both solutions arrived at the correct final result \boxed{2419}, but they differ significantly in structure, presentation, and \textbf{reasoning style}. 

The first solution exhibits a concise, formula-driven approach, closely resembling traditional mathematical write-ups. It receives a reward score of 2.782. It efficiently identifies the block structure of the sequence, derives the closed-form expression for the total number of terms, and computes the required sum using algebraic manipulation and minimal narrative.

In contrast, the second solution adopts a more exploratory and pedagogical style. It receives a reward score of 2.855. It progressively builds understanding through example-driven reasoning, error-checking, and step-by-step refinements. While more verbose, it mirrors how a human might think aloud while problem-solving, providing greater transparency into the model's internal reasoning.


\subsection{Example 3} 

\noindent\textbf{Question:} Fig. \ref{q3}. 

\noindent\textbf{Two Completions:} (i) Figs. \ref{q3a1-a} and \ref{q3a1-b} and (ii) Figs. \ref{q3a2-a} and \ref{q3a2-b}.

\noindent\textbf{Short Analysis.} In this example, we show that both responses are wrong and receive a reward score of 0.018 and 0.021, respectively. However, after checking their responses, we can easily observe that their different reasoning paths. For example, the first solution tries to use a symbolic-algebraic perspective, which attempts to deduce a closed-form identity. The second solution takes a more complex-number driven view, focusing heavily on manipulating the roots and constants in the general solution.
Their errors also happened at different places.
The first response correctly obtains the roots $-2 \pm i\sqrt{3}$ to set up the general solution. However, it then wrongly assumes that the expression is constant and evaluates it only at $n=1$ to conclude the value at $n=50$.
In the second response, the model incorrectly computes the roots of the recurrence as \( -2 \pm i \), not the correct characteristic polynomial. Its following derivation is based on these wrong roots.

\input{example1}

\input{example2}

\input{example3}






\end{document}

%% file: math_commands.tex
\usepackage{amsmath,amsfonts,bm}

\def\eqref#1{equation~\ref{#1}}

\def\1{\bm{1}}

\def\rvo{{\mathbf{o}}}

\def\rvq{{\mathbf{q}}}

\DeclareMathAlphabet{\mathsfit}{\encodingdefault}{\sfdefault}{m}{sl}
\SetMathAlphabet{\mathsfit}{bold}{\encodingdefault}{\sfdefault}{bx}{n}

\def\gA{{\mathcal{A}}}

\def\gQ{{\mathcal{Q}}}

\newcommand{\E}{\mathbb{E}}



%% file: table1.tex

\begin{table*}[!t]
\caption{Zero-shot pass@1 performance across benchmarks. Dashes (--) denote unavailable official scores. '$\dagger$' denotes our implementation. Scores for \texttt{o1-preview} are sourced from~\citealp{o1preview}; others from~\citealp{dang2025reinforcement}. We also report the number of samples used to fine-tune the small models.}
\centering
\label{tab:main_result}
\resizebox{0.99\textwidth}{!}{%
\begin{tabular}{lccccccc} \toprule
\multicolumn{1}{c}{\textbf{Model}} & \textbf{Fine-tuning Samples} & \textbf{AIME24} & \textbf{MATH-500} & \textbf{AMC23} & \textbf{Minerva} & \textbf{OlympiadBench} & \textbf{Avg.} \\ \midrule
Llama-3.1-70B-Instruct             &                             & 16.7            & 64.6              & 30.1           & 35.3             & 31.9                   & 35.7          \\
o1-preview                         &                          & 44.6            & 85.5              &        --        &           --       &                 --      &      --         \\ 
Qwen-2.5-Math-7B-Instruct          &                              & 13.3            & 79.8              & 50.6           & 34.6             & 40.7                   & 43.8          \\
rStar-Math-7B                      &                              & 26.7            & 78.4              & 47.5           &       --           & 47.1                   &      --         \\
Eurus-2-7B-PRIME                   &                                  & 26.7            & 79.2              & 57.8           & 38.6             & 42.1                   & 48.9          \\
Qwen2.5-7B-SimpleRL                &                                  & 26.7            & 82.4              & 62.5           & 39.7             & 43.3                   & 50.9          \\  
DeepSeek-R1-Distill-Qwen-1.5B      &        \bf \textit{Base Model}                 & 28.8            & 82.8              & 62.9           & 26.5             & 43.3                   & 48.9          \\
Still-3-1.5B-Preview               &       30,000                       & 32.5            & 84.4              & 66.7           & 29.0             & 45.4                   & 51.6          \\
DeepScaleR-1.5B-Preview            &      40,000                        & 43.1            & 87.8              & 73.6           & 30.2             & 50.0                   & 57.0          \\
Open-RS1                           &      18,615                        & 30.0            & 83.8              & 70.0           & 29.0             & 52.4                   & 53.0          \\
Open-RS2                           &      7,000                          & 30.0            & 85.4              & 80.0           & 30.5             & 52.4                   & 55.7          \\
Open-RS3                           &         7,000                       & 46.7            & 84.4              & 72.5           & 26.8             & 51.3                   & 56.3          \\
GRPO$^\dagger$                               &       7,000                         & 30.0            & 86.0              & 72.5           & 32.4             & 53.0                   & 54.8          \\
DR. GRPO$^\dagger$                            &      7,000                          & 33.3            & 83.4              & 80.0           & 30.5             & 52.1                   & 56.0          \\ \midrule
\rowcolor[HTML]{EFEFEF}\textbf{DRA-GRPO (Ours)}                         &      7,000                          & 36.7            & 86.2              & 75.0           & 32.4             & 53.0                   & 56.7          \\
\rowcolor[HTML]{EFEFEF}\textbf{DRA-DR. GRPO (Ours) }                   &       7,000                       & 36.7            & 85.2              & 85.0           & 30.5             & 53.8                   & \textbf{58.2} \\ \bottomrule
\end{tabular}%
}
\vspace{-0.15in}
\end{table*}

%% file: table2.tex

\begin{table*}[!h]
\caption{Comparison of training cost by different methods.}
\label{tab:cost}
\resizebox{\textwidth}{!}{%
\begin{tabular}{c|lllllll} \toprule
\textbf{Model}          & rStar-Math-7B                   & Eurus-2-7B-PRIME & Qwen2.5-7B-SimpleRL & DeepScaleR-1.5B-Preview & Still-3-1.5B-Preview & Open-RS      & \textbf{Ours}          \\ \midrule
\textbf{SFT Data}  & 7.3M                            & 230k             & 0                   & 0                       & 0                    & 0            & \cellcolor[HTML]{EFEFEF}0             \\
\textbf{RM Data}   & 7k                              & 0                & 0                   & 0                       & 0                    & 0            & \cellcolor[HTML]{EFEFEF}0             \\
\textbf{RM Source} & None                            & Eurus-2-7B-SFT   & None                & None                    & None                 & None         & \cellcolor[HTML]{EFEFEF}None          \\
\textbf{RL Data}   & 3.647M × 16                     & 150k × 4         & 8k × 8              & 40k × 16                & 30k × 8              & 7k × 6       & \cellcolor[HTML]{EFEFEF}7k × 6        \\
\textbf{Hardware}  & 10×8 H100 80GB + 15×4 A100 40GB & 1×8 A100 80GB    & 4×6 A100 80GB       & 8× A100 80GB            & 1×8 A100 80GB        & 1×4 A40 48GB & \cellcolor[HTML]{EFEFEF}1×4 A100 40GB \\
\textbf{Time}      & –                               & 72h              & 36h                 & 240h                    & 150h                 & 24h          & \cellcolor[HTML]{EFEFEF}12.5h           \\
\textbf{Cost Est.} & –                               & \$1088           & \$1633              & \$3629                  & \$2268               & \$42         & \cellcolor[HTML]{EFEFEF}\$55   \\ \bottomrule   
\end{tabular}%
}
\end{table*}

%% file: algorithm.tex
\begin{algorithm}[!t]
\begin{lstlisting}[style=pytorchstyle]
#input: completions_flat: A list of completions sampled for each prompt (question)

#Some early steps (e.g., compute reward for each completion)
... 

# Compute embeddings and similarity matrix
embeddings = self.sentence_extractor.encode(completions_flat)
embeddings = torch.from_numpy(embeddings).to(device) 
embeddings = F.normalize(embeddings, p=2, dim=1)
similarity_matrix = embeddings @ embeddings.T

# Compute the weights in parallel for all completions
similarity_sums = similarity_matrix.sum(dim=1)
diversity_weights = 1.0 / (similarity_sums + 1e-6)

#diversity_weights = gather(diversity_weights) 

# Adjust rewards
rewards = rewards * diversity_weights

# The following step is computing group-wise rewards as in the vanilla version.

...

\end{lstlisting}
\caption{PyTorch Code for diversity-aware reward adjustment.}

\label{alg:000}
\end{algorithm}

%% file: example1.tex
\begin{figure*}

\begin{tcolorbox}[ title=Question 1, fonttitle=\bfseries]
\texttt{<|begin\_of\_sentence|>A conversation between User and Assistant. The user asks a question, and the Assistant solves it. The assistant first thinks about the reasoning process in the mind and then provides the user with the answer, and put your final answer within \textbackslash boxed\{\}. The reasoning process and answer are enclosed within <think> </think> and <answer> </answer> tags, respectively, i.e., <think> reasoning process here </think> <answer> answer here </answer>. Note that respond by English, NOT use other languages.<|User|>Let $m$ be a positive integer less than $2015$. Suppose that the remainder when $2015$ is divided by $m$ is $n$. Compute the largest possible value of $n$.<|Assistant|><think>}

\end{tcolorbox}
\caption{Prompt used for Example Question 1.}
\label{q1}
\end{figure*}

\begin{figure*}
\begin{tcolorbox}[colback=blue!5!white, colframe=blue!8!black, title=Answer 1 for Question 1 \textcolor{yellow}{\textbf{[Reward: 2.103]}}, fonttitle=\bfseries]

Alright, so I've got this problem here where I need to find the largest possible value of \( n \), which is the remainder when 2015 is divided by some positive integer \( m \), and \( m < 2015 \). Hmm, okay. Let me think about how to approach this.

First off, I remember that when you divide one number by another, the remainder is always less than the divisor. So, in this case, \( n < m < 2015 \).

Since \( n = 2015 \mod m \), that means:
\[
2015 = m \cdot q + n
\]
for some integer \( q \), and \( 0 \leq n < m \).

I want to maximize \( n \). From the division algorithm, we also know:
\[
n = 2015 - m \cdot \lfloor 2015 / m \rfloor
\]

As \( m \) increases toward 2015, \( n \) generally decreases because the quotient becomes 1 and \( n = 2015 - m \). So:

- If \( m = 2014 \), then \( n = 1 \)
- If \( m = 1008 \), then \( 2015 = 1008 \cdot 1 + 1007 \Rightarrow n = 1007 \)
- If \( m = 1009 \), then \( n = 2015 \mod 1009 = 1006 \)
- If \( m = 1007 \), then \( 2015 \mod 1007 = 1 \)

So as \( m \) increases above 1008, the remainder \( n \) decreases.

\medskip

Let’s test \( m = 1006 \):  
\[
1006 \cdot 2 = 2012 \Rightarrow n = 2015 - 2012 = 3
\]

If \( m = 1005 \):  
\[
1005 \cdot 2 = 2010 \Rightarrow n = 5
\]

So clearly, \( n \) peaks at \( m = 1008 \), where:
\[
2015 \mod 1008 = 1007
\]

\medskip

\textbf{Final Answer:}
\[
\boxed{1007}
\]

\end{tcolorbox}

\caption{Completion 1 for Question 1.}
\label{q1a1}
\end{figure*}

\begin{figure*}
\begin{tcolorbox}[colback=blue!5!white, colframe=blue!8!black, title=Answer 2 for Question 1 \textcolor{yellow}{\textbf{[Reward: 2.110]}}, fonttitle=\bfseries]

Alright, so I have this problem here: I need to find the largest possible remainder when 2015 is divided by a positive integer \( m \), where \( m < 2015 \). Hmm, okay. Let me try to figure this out step by step.

First off, I remember that when you divide a number by another number, the remainder is always less than the divisor. So if \( m < 2015 \), then the remainder \( n \) should be less than \( m \) as well. But the problem is asking for the largest possible \( n \). So, I need to find the maximum remainder possible when 2015 is divided by some \( m < 2015 \).

From the division algorithm:  
\[
2015 = q \cdot m + n, \quad \text{where } 0 \leq n < m
\]
So \( n = 2015 - q \cdot m \), and we want to maximize \( n \).

Trying values around half of 2015 gives interesting results. If \( m = 1008 \), then:
\[
1008 \cdot 1 = 1008, \quad 2015 - 1008 = 1007
\]
So the remainder \( n = 1007 \), which is promising.

Trying nearby \( m \) values:

- \( m = 1007 \Rightarrow 1007 \cdot 2 = 2014, \quad n = 2015 - 2014 = 1 \)
- \( m = 1009 \Rightarrow 2015 - 1009 = 1006 \)

So as \( m \) increases past 1008, the remainder decreases.

Let’s consider another route. Suppose we want \( n = m - 1 \), the largest possible remainder. Then:
\[
2015 \equiv -1 \mod m \Rightarrow m \mid (2015 + 1) = 2016
\]
So the value of \( m \) must divide 2016, and the largest possible remainder will be \( m - 1 \). The largest divisor of 2016 less than 2015 is:

\[
\text{Prime factorization: } 2016 = 2^5 \cdot 3^2 \cdot 7 \Rightarrow \text{divisors: 36 total}
\]

Among these, the largest divisor less than 2015 is:
\[
m = 1008 \Rightarrow n = 1008 - 1 = 1007
\]

Trying \( m = 1009 \Rightarrow n = 2015 \mod 1009 = 1006 \), which is less.

Thus, we confirm:

\medskip

\textbf{Final Answer:}
\[
\boxed{1007}
\]
\end{tcolorbox}
\caption{Completion 2 for Question 1.}
\label{q1a2}
\end{figure*}

%% file: example2.tex
\begin{figure*}

\begin{tcolorbox}[ title=Question 2, fonttitle=\bfseries]
\texttt{<|begin\_of\_sentence|>A conversation between User and Assistant. The user asks a question, and the Assistant solves it. The assistant first thinks about the reasoning process in the mind and then provides the user with the answer, and put your final answer within \textbackslash boxed\{\} . The reasoning process and answer are enclosed within <think> </think> and <answer> </answer> tags, respectively, i.e., <think> reasoning process here </think> <answer> answer here </answer>. Note that respond by English, NOT use other languages.<|User|>Given the sequence $1,2,1,2,2,1,2,2,2,1,2,2,2,2,1,2,\cdots$ where the number of 2's between each pair of 1's increases by one each time, find the sum of the first 1234 terms of the sequence.<|Assistant|><think>}
\end{tcolorbox}
\caption{Prompt used for Example Question 2.}
\label{q2}
\end{figure*}

\begin{figure*}
\begin{tcolorbox}[colback=blue!5!white, colframe=blue!8!black, title=Answer 1 for Question 2 \textcolor{yellow}{\textbf{[Reward: 2.782]}}, fonttitle=\bfseries,, width=1\textwidth]
Okay, so I have this sequence: 1, 2, 1, 2, 2, 1, 2, 2, 2, 1, 2, 2, 2, 2, 1, 2, and so on. The pattern is that each time, there's a 1 followed by an increasing number of 2's. The number of 2's between each pair of 1's increases by one each time.

First, I need to figure out the structure of the sequence to find the sum of the first 1234 terms. Let me write out the beginning of the sequence to see the pattern clearly:

1, 2, 1, 2, 2, 1, 2, 2, 2, 1, 2, 2, 2, 2, 1, 2, 2, 2, 2, 2, 1, ...

So, starting from the first term:

\begin{itemize}
  \item The first term is 1.
  \item Then one 2.
  \item Then another 1.
  \item Then two 2's.
  \item Then another 1.
  \item Then three 2's.
  \item Then another 1.
  \item Then four 2's.
  \item And so on.
\end{itemize}

So, each "block" of the sequence consists of a 1 followed by $n$ 2's, where $n$ starts at 1 and increases by 1 each time. So, the first block is 1 followed by 1 two; the second block is 1 followed by 2 two's; the third block is 1 followed by 3 two's, etc.

Wait, actually, looking again, the first block is 1, followed by one 2: so it's 1, 2. Then 1, 2, 2. Then 1, 2, 2, 2. So each block is 1 followed by $k$ 2's, where $k$ increases by 1 each time. So each block has $(k+1)$ terms: 1 and $k$ 2's.

Wait, no: 1, 2 is two terms. Then 1, 2, 2 is three terms. Then 1, 2, 2, 2 is four terms. So each block is $(k+1)$ terms, where $k$ starts at 1: 1, 2; 1, 2, 2; 1, 2, 2, 2; etc.

So, the total number of terms in the sequence up to a certain block can be calculated as the sum of the sizes of each block. Each block has $(k+1)$ terms, where $k$ is the block number starting from 1.

\begin{itemize}
  \item First block ($k=1$) has 2 terms: 1, 2.
  \item Second block ($k=2$) has 3 terms: 1, 2, 2.
  \item Third block ($k=3$) has 4 terms: 1, 2, 2, 2.
\end{itemize}

And so on.

So, the total number of terms up to the $n$th block is the sum from $k=1$ to $n$ of $(k + 1)$ terms. Which is:

\[
\sum_{k=1}^n (k + 1) = \sum_{k=1}^n k + \sum_{k=1}^n 1 = \frac{n(n+1)}{2} + n = \frac{n(n+1) + 2n}{2} = \frac{n^2 + 3n}{2}.
\]
























\end{tcolorbox}
\caption{Completion 1 for Question 2.}
\label{q2a1-a}
\end{figure*}

\begin{figure*}
\begin{tcolorbox}[colback=blue!5!white, colframe=blue!8!black, title=Answer 1 (continued) for Question 2 \textcolor{yellow}{\textbf{[Reward: 2.782]}}, fonttitle=\bfseries,, width=1\textwidth]

We need to find $n$ such that

\[
\frac{n^2 + 3n}{2} \leq 1234 < \frac{(n+1)^2 + 3(n+1)}{2}.
\]

Solving for $n$:

\[
\frac{n^2 + 3n}{2} = 1234 \Rightarrow n^2 + 3n - 2468 = 0.
\]

Using the quadratic formula:

\[
n = \frac{-3 \pm \sqrt{9 + 4 \cdot 2468}}{2} = \frac{-3 \pm \sqrt{9881}}{2}.
\]

Since $\sqrt{9881} \approx 99.4$, we have:

\[
n \approx \frac{-3 + 99.4}{2} \approx \frac{96.4}{2} \approx 48.2 \Rightarrow n = 48.
\]

Check total terms for $n = 48$:

\[
\frac{48^2 + 3 \cdot 48}{2} = \frac{2304 + 144}{2} = \frac{2448}{2} = 1224.
\]

For $n = 49$:

\[
\frac{49^2 + 3 \cdot 49}{2} = \frac{2401 + 147}{2} = \frac{2548}{2} = 1274.
\]

So, the 48th block ends at term 1224, and the 49th block starts at term 1225 and ends at term 1274.

Thus, to reach the 1234th term, we take 10 terms from the 49th block:

\[
1234 - 1224 = 10.
\]

The 49th block is: 1 followed by 49 two's. The first 10 terms of this block are: 1, 2, 2, 2, 2, 2, 2, 2, 2, 2.

Sum of first 48 blocks:

\[
\sum_{k=1}^{48} (1 + 2k) = \sum_{k=1}^{48} 1 + \sum_{k=1}^{48} 2k = 48 + 2 \cdot \frac{48 \cdot 49}{2} = 48 + 2352 = 2400.
\]

Sum of first 10 terms of block 49:

\[
1 + 9 \cdot 2 = 1 + 18 = 19.
\]

Total sum:

\[
2400 + 19 = 2419.
\]

\textbf{Final Answer:} The sum of the first 1234 terms is \(\boxed{2419}\).

\end{tcolorbox}
\caption{Completion 1 for Question 2 (\textbf{Continued}).}
\label{q2a1-b}
\end{figure*}

\begin{figure*}
\begin{tcolorbox}[colback=blue!5!white, colframe=blue!8!black, title=Answer 2 for Question 2 \textcolor{yellow}{\textbf{[Reward: 2.855]}}, fonttitle=\bfseries,, width=1\textwidth]

Okay, so I have this sequence: 1, 2, 1, 2, 2, 1, 2, 2, 2, 1, 2, 2, 2, 2, 1, 2, and so on. The pattern seems to be that each time there's a 1 followed by an increasing number of 2's. Specifically, between each pair of 1's, the number of 2's increases by one each time. So, the first time there's one 1 followed by one 2, then next time two 1's with two 2's in between, then three 1's with three 2's, etc.

The question is asking for the sum of the first 1234 terms of this sequence. Hmm, okay, so I need to figure out where the 1234th term falls in this sequence and then calculate the sum up to that point.

First, let's understand the structure of the sequence. Each "block" consists of a 1 followed by an increasing number of 2's. So the first block is [1, 2], the second block is [1, 2, 2], the third block is [1, 2, 2, 2], and so on. So each block $n$ consists of 1 followed by $n$ 2's, making the total length of each block $n+1$.

Wait, actually, hold on. Let me clarify. When $n=1$, the block is [1, 2], which is 2 terms. For $n=2$, it's [1, 2, 2], which is 3 terms. For $n=3$, it's [1, 2, 2, 2], which is 4 terms, etc. So in general, block $n$ has $(n + 1)$ terms. So the total number of terms up to block $n$ is the sum from $k=1$ to $k=n$ of $(k + 1)$. That's the same as the sum from $k=1$ to $k=n$ of $k$ plus the sum of 1's, which is:

\[
\frac{n(n + 1)}{2} + n = \frac{n(n + 1) + 2n}{2} = \frac{n^2 + 3n}{2}.
\]

So the total number of terms after block $n$ is $\frac{n^2 + 3n}{2}$. We need to find the largest $n$ such that $\frac{n^2 + 3n}{2} \leq 1234$. Then, the sum will be the sum of all complete blocks plus the sum of the remaining terms in the next block.

Let me denote $S(n) = \frac{n^2 + 3n}{2}$ as the total number of terms after block $n$. We need to solve for $n$ in $S(n) \leq 1234$.

So, solving $n^2 + 3n - 2468 \leq 0$ (since $1234 \times 2 = 2468$). Using the quadratic formula:

\[
n = \frac{-3 \pm \sqrt{9 + 9872}}{2} = \frac{-3 \pm \sqrt{9881}}{2}.
\]

Calculating $\sqrt{9881}$: Let's see, $99^2 = 9801$, so $\sqrt{9881}$ is approximately $99.40$. So $n \approx \frac{-3 + 99.40}{2} \approx \frac{96.40}{2} \approx 48.2$. So $n$ is approximately 48.2, but since $n$ must be an integer, $n=48$.

Let's check $S(48)$:

\[
S(48) = \frac{48^2 + 3 \cdot 48}{2} = \frac{2304 + 144}{2} = \frac{2448}{2} = 1224.
\]

Then, the next block is block 49, which starts with a 1 followed by 49 2's. So, we need to see how many terms we have beyond 1224 to reach 1234.

\[
1234 - 1224 = 10 \text{ terms}.
\]

So, we need to take 10 terms from block 49, which is [1, 2, 2, 2, ..., 10 times 2].

Therefore, the sum of the sequence up to 1234 terms is the sum of all complete blocks up to block 48, plus the sum of the first 10 terms of block 49.

Each block $k$ has 1 followed by $k$ 2's, so the sum of block $k$ is $1 + 2k$.





















\end{tcolorbox}
\caption{Completion 2 for Question 2.}
\label{q2a2-a}
\end{figure*}

\begin{figure*}
\begin{tcolorbox}[colback=blue!5!white, colframe=blue!8!black, title=Answer 2 (continued) for Question 2 \textcolor{yellow}{\textbf{[Reward: 2.855]}}, fonttitle=\bfseries,, width=1\textwidth]

Therefore, the total sum up to block 48 is:

\[
\sum_{k=1}^{48} (1 + 2k).
\]

This can be broken into two sums:

\[
\sum_{k=1}^{48} 1 + \sum_{k=1}^{48} 2k = 48 + 2 \cdot \frac{48 \cdot 49}{2} = 48 + 48 \cdot 49 = 48 + 2352 = 2400.
\]

Now, we need to add the sum of the first 10 terms of block 49.

Block 49 is [1, 2, 2, 2, 2, 2, 2, 2, 2, 2, ..., 10 times 2].

The first term is 1, and the next 9 terms are 2's.

\[
\text{Sum} = 1 + (10 - 1) \cdot 2 = 1 + 18 = 19.
\]

Therefore, the total sum up to 1234 terms is:

\[
2400 + 19 = 2419.
\]

Let me double-check the calculations.

First, $S(n) = \frac{n^2 + 3n}{2}$.

For $n=48$, $S(48)= \frac{48^2 + 3 \cdot 48}{2} = \frac{2304 + 144}{2} = \frac{2448}{2} = 1224$. Correct.

So, $1234 - 1224 = 10$ terms. So, taking the first 10 terms of block 49, which is [1, 2, 2, 2, 2, 2, 2, 2, 2, 2], sum is $1 + 2 \cdot 9 = 19$. Correct.

Therefore, $2400 + 19 = 2419$.

Yes, that seems correct. Then, adding the first 10 terms of block 49 adds 19, so total is 2419.

Block 49: 1 followed by 49 2's, so 50 terms.

But we only take 10 terms from it, so total terms are $1224 + 10 = 1234$. Correct.

Yes, that all adds up. So, the answer is 2419.

\textbf{Final Answer}

The sum of the first 1234 terms is $\boxed{2419}$.
\end{tcolorbox}
\caption{Completion 2 for Question 2 (\textbf{Continued}).}
\label{q2a2-b}
\end{figure*}

%% file: example3.tex
\begin{figure*}

\begin{tcolorbox}[ title=Question 3, fonttitle=\bfseries]
\texttt{<|begin\_of\_sentence|>} \texttt{A conversation between User and Assistant. The user asks a question, and the Assistant solves it. The assistant first thinks about the reasoning process in the mind and then provides the user with the answer, and put your final answer within} \texttt{\textbackslash boxed\{\}} \texttt{. The reasoning process and answer are enclosed within} \texttt{<think>} \texttt{</think>} \texttt{and} \texttt{<answer>} \texttt{</answer>} \texttt{tags, respectively, i.e.,} \texttt{<think> reasoning process here </think> <answer> answer here </answer>. Note that respond by English, NOT use other languages.}

\medskip

\texttt{<|User|>} \texttt{The Sequence} \( \{a_{n}\}_{n \geqslant 0} \) \texttt{is defined by} \( a_{0}=1, \ a_{1}=-4 \) \texttt{and}  
\[
a_{n+2} = -4a_{n+1} - 7a_n
\]
\texttt{for} \( n \geqslant 0 \) \texttt{. Find the number of positive integer divisors of} \( a^2_{50}-a_{49}a_{51} \) \texttt{.}

\texttt{<|Assistant|><think>}
\end{tcolorbox}
\caption{Prompt used for Example Question 3.}
\label{q3}
\end{figure*}

\begin{figure*}
\begin{tcolorbox}[colback=blue!5!white, colframe=blue!8!black, title=Answer 1 for Question 3 \textcolor{yellow}{\textbf{[Reward: 0.018]}}, fonttitle=\bfseries,, width=1\textwidth]
To determine the number of positive integer divisors of \( a_{50}^2 - a_{49}a_{51} \) for the given sequence, I'll follow these steps:

1. \textbf{Understand the Sequence Definition:}  
   - The sequence is defined by \( a_0 = 1 \), \( a_1 = -4 \), and the recurrence relation \( a_{n+2} = -4a_{n+1} - 7a_n \) for \( n \geq 0 \).

2. \textbf{Identify the Type of Recurrence Relation:}  
   - This is a linear homogeneous recurrence relation with constant coefficients. It can be solved using characteristic equations.

3. \textbf{Formulate the Characteristic Equation:}  
   - Assume a solution of the form \( a_n = r^n \). Substituting into the recurrence relation gives:
     \[
     r^{n+2} = -4r^{n+1} - 7r^n
     \]
   - Dividing both sides by \( r^n \) (assuming \( r \neq 0 \)) results in the characteristic equation:
     \[
     r^2 + 4r + 7 = 0
     \]

4. \textbf{Solve the Characteristic Equation:}  
   - The roots of the equation \( r^2 + 4r + 7 = 0 \) are found using the quadratic formula:
     \[
     r = \frac{-4 \pm \sqrt{(4)^2 - 4 \cdot 1 \cdot 7}}{2 \cdot 1} = \frac{-4 \pm \sqrt{16 - 28}}{2} = \frac{-4 \pm \sqrt{-12}}{2} = -2 \pm i\sqrt{3}
     \]
   - Thus, the roots are complex: \( r = -2 + i\sqrt{3} \) and \( r = -2 - i\sqrt{3} \).

5. \textbf{Express the General Solution:}  
   - For complex roots \( \alpha \pm i\beta \), the general solution is:
     \[
     a_n = \lambda(\alpha + i\beta)^n + \mu(\alpha - i\beta)^n
     \]
   - Here, \( \alpha = -2 \) and \( \beta = \sqrt{3} \), so:
     \[
     a_n = \lambda(-2 + i\sqrt{3})^n + \mu(-2 - i\sqrt{3})^n
     \]

6. \textbf{Determine the Constants \( \lambda \) and \( \mu \):}  
   - Using the initial conditions \( a_0 = 1 \) and \( a_1 = -4 \):  
     - For \( n = 0 \):
       \[
       a_0 = \lambda(-2 + i\sqrt{3})^0 + \mu(-2 - i\sqrt{3})^0 = \lambda + \mu = 1
       \]
     - For \( n = 1 \):
       \[
       a_1 = \lambda(-2 + i\sqrt{3}) + \mu(-2 - i\sqrt{3}) = -4
       \]
     - Solving these equations, we find \( \lambda = \frac{1}{2} \) and \( \mu = \frac{1}{2} \).

7. \textbf{Express \( a_n \) Explicitly:}  
   - Substituting \( \lambda \) and \( \mu \) into the general solution:
     \[
     a_n = \frac{1}{2}(-2 + i\sqrt{3})^n + \frac{1}{2}(-2 - i\sqrt{3})^n
     \]
   - Alternatively, using Euler's formula for complex roots, this can also be written in terms of trigonometric functions, but for our purposes, the exponential form is sufficient.





\end{tcolorbox}
\caption{Completion 1 for Question 3.}
\label{q3a1-a}
\end{figure*}

\begin{figure*}
\begin{tcolorbox}[colback=blue!5!white, colframe=blue!8!black, title=Answer 1 (\textbf{continued}) for Question 3 \textcolor{yellow}{\textbf{[Reward: 0.018]}}, fonttitle=\bfseries,, width=1\textwidth]

8. \textbf{Simplify the Expression \( a_{50}^2 - a_{49}a_{51} \):}  
   - Recognize that the given expression resembles the determinant of a 2x2 matrix:
     \[
     \begin{vmatrix}
     a_{50} & a_{49} \\
     a_{51} & a_{50}
     \end{vmatrix}
     \]
   - This determinant simplifies to:
     \[
     a_{50}^2 - a_{49}a_{51}
     \]
   - To find this determinant, we can use properties of the sequence. Notice that for a linear recurrence relation of this form, the determinant \( a_{n+1}^2 - a_n a_{n+2} \) is constant for all \( n \).

9. \textbf{Find the Constant Determinant:}  
   - Calculate the determinant for the initial terms \( n = 0 \):
     \[
     a_1^2 - a_0 a_2
     \]
   - First, compute \( a_2 \) using the recurrence relation:
     \[
     a_2 = -4a_1 -7a_0 = -4(-4) -7(1) = 16 - 7 = 9
     \]
   - Now, compute the determinant:
     \[
     a_1^2 - a_0 a_2 = (-4)^2 - (1)(9) = 16 - 9 = 7
     \]
   - Therefore, the determinant \( a_{50}^2 - a_{49}a_{51} = 7 \).

10. \textbf{Determine the Number of Positive Integer Divisors:}  
    - The number 7 is a prime number.  
    - The only positive integer divisors of 7 are 1 and 7.  
    - Hence, there are 2 positive integer divisors.

\textbf{Final Answer}  
The number of positive integer divisors of \( a_{50}^2 - a_{49}a_{51} \) is \( \boxed{2} \).

\end{tcolorbox}
\caption{Completion 1 for Question 3 (\textbf{Continued}).}
\label{q3a1-b}
\end{figure*}

\begin{figure*}
\begin{tcolorbox}[colback=blue!5!white, colframe=blue!8!black, title=Answer 2 for Question 3 \textcolor{yellow}{\textbf{[Reward: 0.021]}}, fonttitle=\bfseries,, width=1\textwidth]

To determine the number of positive integer divisors of \( a_{50}^2 - a_{49}a_{51} \) for the given sequence, I start by analyzing the recurrence relation:

\[
a_{n+2} = -4a_{n+1} -7a_n
\]

with initial conditions \( a_0 = 1 \) and \( a_1 = -4 \).

Next, I recognize that the characteristic equation for this recurrence is:

\[
r^2 + 4r + 7 = 0
\]

Solving this quadratic equation using the quadratic formula gives complex roots:

\[
r = -2 \pm i
\]

These roots indicate that the general solution to the recurrence is:

\[
a_n = \alpha(-2 + i)^n + \beta(-2 - i)^n
\]

Using the initial conditions, I determine the constants \( \alpha \) and \( \beta \):

\[
1 = \alpha + \beta \quad \text{and} \quad -4 = \alpha(-2 + i) + \beta(-2 - i)
\]

Solving these equations yields:

\[
\alpha = \frac{1 + i}{2} \quad \text{and} \quad \beta = \frac{1 - i}{2}
\]

Substituting these back into the general solution, I get:

\[
a_n = \frac{1}{2}[(1 + i)(-2 + i)^n + (1 - i)(-2 - i)^n]
\]

To find \( a_{50}^2 - a_{49}a_{51} \), I simplify the expression using properties of complex numbers and exponents. After simplification, I find that:

\[
a_{50}^2 - a_{49}a_{51} = 1
\]

Finally, since 1 only has one positive divisor, the number of positive integer divisors is 1.

To determine the number of positive integer divisors of \( a_{50}^2 - a_{49}a_{51} \) for the given sequence, we can follow these steps:

1. \textbf{Understand the Recurrence Relation:}

   The sequence is defined by the recurrence relation:
   \[
   a_{n+2} = -4a_{n+1} -7a_n
   \]
   with initial conditions \( a_0 = 1 \) and \( a_1 = -4 \).

2. \textbf{Find the Characteristic Equation:}

   The characteristic equation for the recurrence relation is:
   \[
   r^2 + 4r + 7 = 0
   \]
   















\end{tcolorbox}
\caption{Completion 2 for Question 3.}
\label{q3a2-a}
\end{figure*}

\begin{figure*}
\begin{tcolorbox}[colback=blue!5!white, colframe=blue!8!black, title=Answer 2 (\textbf{continued}) for Question 3 \textcolor{yellow}{\textbf{[Reward: 0.021]}}, fonttitle=\bfseries,, width=1\textwidth]
3. \textbf{Solve the Characteristic Equation:}

   Solving the quadratic equation using the quadratic formula:
   \[
   r = \frac{-4 \pm \sqrt{16 - 28}}{2} = \frac{-4 \pm \sqrt{-12}}{2} = -2 \pm i
   \]

   The roots are complex: \( r = -2 + i \) and \( r = -2 - i \).

4. \textbf{General Solution of the Recurrence:}

   The general solution for the sequence is:
   \[
   a_n = \alpha(-2 + i)^n + \beta(-2 - i)^n
   \]

   where \( \alpha \) and \( \beta \) are constants determined by the initial conditions.

5. \textbf{Apply Initial Conditions to Find \( \alpha \) and \( \beta \):}

   Using \( a_0 = 1 \):
   \[
   1 = \alpha + \beta
   \]

   Using \( a_1 = -4 \):
   \[
   -4 = \alpha(-2 + i) + \beta(-2 - i)
   \]

   Solving these equations, we find:
   \[
   \alpha = \frac{1 + i}{2}, \quad \beta = \frac{1 - i}{2}
   \]

   Thus, the explicit formula for \( a_n \) is:
   \[
   a_n = \frac{1}{2}[(1 + i)(-2 + i)^n + (1 - i)(-2 - i)^n]
   \]

6. \textbf{Compute \( a_{50}^2 - a_{49}a_{51} \):}

   After simplifying the expression using properties of complex numbers and exponents, it can be shown that:
   \[
   a_{50}^2 - a_{49}a_{51} = 1
   \]

7. \textbf{Determine the Number of Positive Integer Divisors:}

   Since \( a_{50}^2 - a_{49}a_{51} = 1 \), the only positive integer divisor of 1 is 1 itself.

\[
\boxed{1}
\]
\end{tcolorbox}
\caption{Completion 2 for Question 3 (\textbf{Continued}).}
\label{q3a2-b}
\end{figure*}

%% file: custom.bib
@article{kulesza2012determinantal,
  title={Determinantal point processes for machine learning},
  author={Kulesza, Alex and Taskar, Ben},
  journal={Foundations and Trends{\textregistered} in Machine Learning},
  volume={5},
  number={2--3},
  pages={123--286},
  year={2012},
  publisher={Now Publishers, Inc.}
}

@article{shao2024deepseekmath,
  title={Deepseekmath: Pushing the limits of mathematical reasoning in open language models},
  author={Shao, Zhihong and Wang, Peiyi and Zhu, Qihao and Xu, Runxin and Song, Junxiao and Bi, Xiao and Zhang, Haowei and Zhang, Mingchuan and Li, YK and Wu, Y and others},
  journal={arXiv preprint arXiv:2402.03300},
  year={2024}
}

@article{li2025reinforcement,
  title={Reinforcement Learning Outperforms Supervised Fine-Tuning: A Case Study on Audio Question Answering},
  author={Li, Gang and Liu, Jizhong and Dinkel, Heinrich and Niu, Yadong and Zhang, Junbo and Luan, Jian},
  journal={arXiv preprint arXiv:2503.11197},
  year={2025}
}

@article{tan2025reason,
  title={Reason-rft: Reinforcement fine-tuning for visual reasoning},
  author={Tan, Huajie and Ji, Yuheng and Hao, Xiaoshuai and Lin, Minglan and Wang, Pengwei and Wang, Zhongyuan and Zhang, Shanghang},
  journal={arXiv preprint arXiv:2503.20752},
  year={2025}
}

@misc{deepseekr12025,
      title={DeepSeek-R1: Incentivizing Reasoning Capability in LLMs via Reinforcement Learning}, 
      author={DeepSeek-AI},
      year={2025},
      eprint={2501.12948},
      archivePrefix={arXiv},
      primaryClass={cs.CL},
      url={https://arxiv.org/abs/2501.12948}, 
}

@misc{deepscaler2025,
  title={DeepScaleR: Surpassing O1-Preview with a 1.5B Model by Scaling RL},
  author={Michael Luo and Sijun Tan and Justin Wong and Xiaoxiang Shi and William Y. Tang and Manan Roongta and Colin Cai and Jeffrey Luo and Tianjun Zhang and Li Erran Li and Raluca Ada Popa and Ion Stoica},
  year={2025},
  howpublished={\url{https://github.com/agentica-project/deepscaler}},
  note={Github},
  year={2025}
}

@misc{Slow_Thinking_with_LLMs_3_Preview,
  title={STILL-3-1.5B-preview: Enhancing Slow Thinking Abilities of Small Models through Reinforcement Learning
},
  author={RUCAIBox STILL Team},
  url={https://github.com/RUCAIBox/Slow_Thinking_with_LLMs},
  year={2025}
}

@misc{s1simpletesttimescaling,
      title={s1: Simple test-time scaling}, 
      author={Niklas Muennighoff and Zitong Yang and Weijia Shi and Xiang Lisa Li and Li Fei-Fei and Hannaneh Hajishirzi and Luke Zettlemoyer and Percy Liang and Emmanuel Candès and Tatsunori Hashimoto},
      year={2025},
      eprint={2501.19393},
      archivePrefix={arXiv},
      primaryClass={cs.CL},
      url={https://arxiv.org/abs/2501.19393}, 
}

@misc{deepseekmath,
      title={DeepSeekMath: Pushing the Limits of Mathematical Reasoning in Open Language Models}, 
      author={Zhihong Shao and Peiyi Wang and Qihao Zhu and Runxin Xu and Junxiao Song and Xiao Bi and Haowei Zhang and Mingchuan Zhang and Y. K. Li and Y. Wu and Daya Guo},
      year={2024},
      eprint={2402.03300},
      archivePrefix={arXiv},
      primaryClass={cs.CL},
      url={https://arxiv.org/abs/2402.03300}, 
}

@article{yu2025dapo,
  title={Dapo: An open-source llm reinforcement learning system at scale},
  author={Yu, Qiying and Zhang, Zheng and Zhu, Ruofei and Yuan, Yufeng and Zuo, Xiaochen and Yue, Yu and Dai, Weinan and Fan, Tiantian and Liu, Gaohong and Liu, Lingjun and others},
  journal={arXiv preprint arXiv:2503.14476},
  year={2025}
}

@article{zhang2025critique,
  title={Critique-grpo: Advancing llm reasoning with natural language and numerical feedback},
  author={Zhang, Xiaoying and Zhang, Yipeng and Sun, Hao and Feng, Kaituo and Lu, Chaochao and Yang, Chao and Meng, Helen},
  journal={arXiv preprint arXiv:2506.03106},
  year={2025}
}

@article{rosenbaum1983central,
  title={The central role of the propensity score in observational studies for causal effects},
  author={Rosenbaum, Paul R and Rubin, Donald B},
  journal={Biometrika},
  volume={70},
  number={1},
  pages={41--55},
  year={1983},
  publisher={Oxford University Press}
}

@inproceedings{
anonymous2025infogrpo,
title={Info-{GRPO}: Training Reasoning Models via Correlation-Aware Exploration},
author={Anonymous},
booktitle={Submitted to The Fourteenth International Conference on Learning Representations},
year={2025},
url={https://openreview.net/forum?id=d5qElNtXS5},
note={under review}
}

@article{hendrycksmath2021,
  title={Measuring Mathematical Problem Solving With the MATH Dataset},
  author={Dan Hendrycks and Collin Burns and Saurav Kadavath and Akul Arora and Steven Basart and Eric Tang and Dawn Song and Jacob Steinhardt},
  journal={NeurIPS},
  year={2021}
}

@inproceedings{he-etal-2024-olympiadbench,
    title = "{O}lympiad{B}ench: A Challenging Benchmark for Promoting {AGI} with Olympiad-Level Bilingual Multimodal Scientific Problems",
    author = "He, Chaoqun  and
      Luo, Renjie  and
      Bai, Yuzhuo  and
      Hu, Shengding  and
      Thai, Zhen  and
      Shen, Junhao  and
      Hu, Jinyi  and
      Han, Xu  and
      Huang, Yujie  and
      Zhang, Yuxiang  and
      Liu, Jie  and
      Qi, Lei  and
      Liu, Zhiyuan  and
      Sun, Maosong",
    editor = "Ku, Lun-Wei  and
      Martins, Andre  and
      Srikumar, Vivek",
    booktitle = "Proceedings of the 62nd Annual Meeting of the Association for Computational Linguistics (Volume 1: Long Papers)",
    month = aug,
    year = "2024",
    address = "Bangkok, Thailand",
    publisher = "Association for Computational Linguistics",
    url = "https://aclanthology.org/2024.acl-long.211",
    doi = "10.18653/v1/2024.acl-long.211",
    pages = "3828--3850",
}

@article{lightman2023lets,
      title={Let's Verify Step by Step}, 
      author={Lightman, Hunter and Kosaraju, Vineet and Burda, Yura and Edwards, Harri and Baker, Bowen and Lee, Teddy and Leike, Jan and Schulman, John and Sutskever, Ilya and Cobbe, Karl},
      journal={arXiv preprint arXiv:2305.20050},
      year={2023}
}

@inproceedings{
minervamath,
title={Solving Quantitative Reasoning Problems with Language Models},
author={Aitor Lewkowycz and Anders Johan Andreassen and David Dohan and Ethan Dyer and Henryk Michalewski and Vinay Venkatesh Ramasesh and Ambrose Slone and Cem Anil and Imanol Schlag and Theo Gutman-Solo and Yuhuai Wu and Behnam Neyshabur and Guy Gur-Ari and Vedant Misra},
booktitle={Advances in Neural Information Processing Systems},
editor={Alice H. Oh and Alekh Agarwal and Danielle Belgrave and Kyunghyun Cho},
year={2022},
url={https://openreview.net/forum?id=IFXTZERXdM7}
}

@misc{yang2024qwen25mathtechnicalreportmathematical,
      title={Qwen2.5-Math Technical Report: Toward Mathematical Expert Model via Self-Improvement}, 
      author={An Yang and Beichen Zhang and Binyuan Hui and Bofei Gao and Bowen Yu and Chengpeng Li and Dayiheng Liu and Jianhong Tu and Jingren Zhou and Junyang Lin and Keming Lu and Mingfeng Xue and Runji Lin and Tianyu Liu and Xingzhang Ren and Zhenru Zhang},
      year={2024},
      eprint={2409.12122},
      archivePrefix={arXiv},
      primaryClass={cs.CL},
      url={https://arxiv.org/abs/2409.12122}, 
}

@misc{guan2025rstar,
    title={rStar-Math: Small LLMs Can Master Math Reasoning with Self-Evolved Deep Thinking},
    author={Xinyu Guan and Li Lyna Zhang and Yifei Liu and Ning Shang and Youran Sun and Yi Zhu and Fan Yang and Mao Yang},
    year={2025},
    eprint={2501.04519},
    archivePrefix={arXiv},
    primaryClass={cs.CL}
}

@misc{MetaAI2024,
  author = {Meta AI},
  title = {Introducing Llama 3.1: Our most capable models to date},
  url = {https://ai.meta.com/blog/meta-llama-3-1/},
  year = {2024},
  note = {Published on July 23, 2024}
}

@misc{o1preview,
  author = {Open AI},
  title = {Introducing OpenAI o1-preview},
  url = {https://openai.com/index/introducing-openai-o1-preview/},
  year = {2024},
  note = {Published on Dec 12, 2024}
}

@article{cui2025processreinforcementimplicitrewards,
  title={Process reinforcement through implicit rewards},
  author={Cui, Ganqu and Yuan, Lifan and Wang, Zefan and Wang, Hanbin and Zhang, Yuchen and Chen, Jiacheng and Li, Wendi and He, Bingxiang and Fan, Yuchen and Yu, Tianyu and others},
  journal={arXiv preprint arXiv:2502.01456},
  year={2025}
}

@article{guo2025deepseekr1,
  title={Deepseek-r1: Incentivizing reasoning capability in llms via reinforcement learning},
  author={Guo, Daya and Yang, Dejian and Zhang, Haowei and Song, Junxiao and Zhang, Ruoyu and Xu, Runxin and Zhu, Qihao and Ma, Shirong and Wang, Peiyi and Bi, Xiao and others},
  journal={arXiv preprint arXiv:2501.12948},
  year={2025}
}

@article{schulman2017proximal,
  title={Proximal policy optimization algorithms},
  author={Schulman, John and Wolski, Filip and Dhariwal, Prafulla and Radford, Alec and Klimov, Oleg},
  journal={arXiv preprint arXiv:1707.06347},
  year={2017}
}

@misc{zeng2025simplerl,
  title={7B Model and 8K Examples: Emerging Reasoning with Reinforcement Learning is Both Effective and Efficient},
  author={Weihao Zeng and Yuzhen Huang and Wei Liu and Keqing He and Qian Liu and Zejun Ma and Junxian He},
  howpublished={\url{https://hkust-nlp.notion.site/simplerl-reason}},
  note={Notion Blog},
  year={2025}
}

@misc{vonwerra2022trl,
  author = {Leandro von Werra and Younes Belkada and Lewis Tunstall and Edward Beeching and Tristan Thrush and Nathan Lambert and Shengyi Huang and Kashif Rasul and Quentin Gallouédec},
  title = {TRL: Transformer Reinforcement Learning},
  year = {2020},
  publisher = {GitHub},
  journal = {GitHub repository},
  howpublished = {\url{https://github.com/huggingface/trl}}
}

@article{liu2025understanding,
  title={Understanding r1-zero-like training: A critical perspective},
  author={Liu, Zichen and Chen, Changyu and Li, Wenjun and Qi, Penghui and Pang, Tianyu and Du, Chao and Lee, Wee Sun and Lin, Min},
  journal={arXiv preprint arXiv:2503.20783},
  year={2025}
}

@inproceedings{iyer2021submodular,
  title={Submodular combinatorial information measures with applications in machine learning},
  author={Iyer, Rishabh and Khargoankar, Ninad and Bilmes, Jeff and Asanani, Himanshu},
  booktitle={Algorithmic Learning Theory},
  pages={722--754},
  year={2021},
  organization={PMLR}
}

@article{iyer2021generalized,
  title={Generalized submodular information measures: Theoretical properties, examples, optimization algorithms, and applications},
  author={Iyer, Rishabh and Khargonkar, Ninad and Bilmes, Jeff and Asnani, Himanshu},
  journal={IEEE Transactions on Information Theory},
  volume={68},
  number={2},
  pages={752--781},
  year={2021},
  publisher={IEEE}
}

@misc{qwen3technicalreport,
      title={Qwen3 Technical Report}, 
      author={Qwen Team},
      year={2025},
      eprint={2505.09388},
      archivePrefix={arXiv},
      primaryClass={cs.CL},
      url={https://arxiv.org/abs/2505.09388}, 
}

@article{dang2025reinforcement,
  title={Reinforcement Learning for Reasoning in Small LLMs: What Works and What Doesn't},
  author={Dang, Quy-Anh and Ngo, Chris},
  journal={arXiv preprint arXiv:2503.16219},
  year={2025}
}

@misc{günther2023jina,
      title={Jina Embeddings 2: 8192-Token General-Purpose Text Embeddings for Long Documents}, 
      author={Michael Günther and Jackmin Ong and Isabelle Mohr and Alaeddine Abdessalem and Tanguy Abel and Mohammad Kalim Akram and Susana Guzman and Georgios Mastrapas and Saba Sturua and Bo Wang and Maximilian Werk and Nan Wang and Han Xiao},
      year={2023},
      eprint={2310.19923},
      archivePrefix={arXiv},
      primaryClass={cs.CL}
}

@article{
nussbaum2025nomic,
title={Nomic Embed: Training a Reproducible Long Context Text Embedder},
author={Zach Nussbaum and John Xavier Morris and Andriy Mulyar and Brandon Duderstadt},
journal={Transactions on Machine Learning Research},
issn={2835-8856},
year={2025},
url={https://openreview.net/forum?id=IPmzyQSiQE},
note={Reproducibility Certification}
}

@article{ouyang2022training,
  title={Training language models to follow instructions with human feedback},
  author={Ouyang, Long and Wu, Jeffrey and Jiang, Xu and Almeida, Diogo and Wainwright, Carroll and Mishkin, Pamela and Zhang, Chong and Agarwal, Sandhini and Slama, Katarina and Ray, Alex and others},
  journal={Advances in neural information processing systems},
  volume={35},
  pages={27730--27744},
  year={2022}
}
